%% file: main.tex
\documentclass[bst/sn-mathphys]{sn-jnl}% Math and Physical Sciences Reference Style
\usepackage[T1]{fontenc}    % use 8-bit T1 fonts
\usepackage{hyperref}       % hyperlinks
\usepackage{url}            % simple URL typesetting
\usepackage{booktabs}       % professional-quality tables
\usepackage{amsfonts}       % blackboard math symbols
\usepackage{nicefrac}       % compact symbols for 1/2, etc.
\usepackage{microtype}      % microtypography
\usepackage{xcolor}         % colors
\usepackage{graphicx}
\usepackage{textcomp}
\usepackage{booktabs}
\usepackage{tabularx}
\usepackage{multirow}
\usepackage{listings}
\usepackage{multicol}
\usepackage{subcaption}
\usepackage{epstopdf}
\usepackage{color, colortbl}
\usepackage{amsthm} 
\usepackage{amssymb}
\usepackage{mathtools}
\usepackage{rotating}
\usepackage{tablefootnote}
\usepackage{colortbl}
\usepackage{xcolor}
\usepackage{fontawesome5}
\usepackage{vcell}
\usepackage{tablefootnote}
\usepackage{comment}
\newcommand{\rev}[1]{\textcolor{black}{#1}}
\jyear{2022}%

\theoremstyle{thmstyleone}%
\theoremstyle{thmstyletwo}%

\theoremstyle{thmstylethree}%

\begin{document}

\title{MEDIC: A Multi-Task Learning Dataset for Disaster Image Classification}
%%Multitask lEarning Dataset Image Classification


\author*[1]{\fnm{Firoj} \sur{Alam}}\email{fialam@hbku.edu.qa}
\author[2]{\fnm{Tanvirul} \sur{Alam}}\email{tanvirul.alam@mail.rit.edu}
\author[3,4]{\fnm{Md. Arid} \sur{Hasan}}\email{arid.cse0325.c@diu.edu.bd}
\author[5]{\fnm{Abul} \sur{Hasnat}}\email{mhasnat@gmail.com}
\author[1]{\fnm{Muhammad} \sur{Imran}}\email{mimran@hbku.edu.qa}
\author[1]{\fnm{Ferda} \sur{Ofli}}\email{fofli@hbku.edu.qa}
\affil*[1]{\orgdiv{Qatar Computing Research Institute}, \orgaddress{\street{HBKU}, \state{Doha}, \country{Qatar}}}
\affil[2]{\orgname{Rochester Institute of Technology}, \orgaddress{\city{Rochester}, \state{Rochester}, \country{USA}}}
\affil[3]{\orgname{Cognitive Insight Limited}, \orgaddress{\city{Dhaka}, \country{Bangladesh}}}
\affil[4]{\orgname{Daffodil International University}, \orgaddress{\city{Dhaka}, \state{Dhaka}, \country{Bangladesh}}}
\affil[5]{\orgname{BLACKBIRD.AI}, \orgaddress{\country{USA}}}

\abstract{
Recent research in disaster informatics demonstrates a practical and important use case of artificial intelligence to save human lives and suffering during natural disasters based on social media contents (text and images). While notable progress has been made using texts, research on exploiting the images remains relatively under-explored. To advance image-based approaches, we propose MEDIC\footnote{Available~at: \url{https://crisisnlp.qcri.org/medic/index.html}}, which is the largest social media image classification dataset for humanitarian response consisting of 71,198 images to address four different tasks in a multi-task learning setup. This is the first dataset of its kind: social media images, disaster response, and multi-task learning research. An important property of this dataset is its high potential to facilitate research on \textit{multi-task learning}, which recently receives much interest from the machine learning community and has shown remarkable results in terms of memory, inference speed, performance, and generalization capability. Therefore, the proposed dataset is an important resource for advancing image-based disaster management and multi-task machine learning research. We experiment with different deep learning architectures and report promising results, which are above the majority baselines for all tasks. Along with the dataset, we also release all relevant scripts.\footnote{\url{https://github.com/firojalam/medic}}
}




\maketitle

\input{sections/introduction}

\input{sections/related_work}

\input{sections/dataset}
\input{sections/experiments}

\input{sections/discussion}
\input{sections/conclusion}

\section*{Declarations}
The authors have no competing interests.

% \begin{itemize}

% \item The authors have no relevant financial or non-financial interests to disclose.
% \item The authors have no competing interests to declare that are relevant to the content of this article.
% \item All authors certify that they have no affiliations with or involvement in any organization or entity with any financial interest or non-financial interest in the subject matter or materials discussed in this manuscript.
% \item The authors have no financial or proprietary interests in any material discussed in this article.
% \end{itemize}
%
%\begin{itemize}
%\item Conflict of interest/Competing interests (check journal-specific guidelines for which heading to use)
%\item Ethics approval 
%\item Consent for publication
%\item Availability of data and materials
%\item Authors' contributions
%\end{itemize}
%
%\noindent
%If any of the sections are not relevant to your manuscript, please include the heading and write `Not applicable' for that section. 
%
%%%===================================================%%
%%% For presentation purpose, we have included        %%
%%% \bigskip command. please ignore this.             %%
%%%===================================================%%
%\bigskip
%\bigskip\noindent
%Springer journals and proceedings: \url{https://www.springer.com/gp/editorial-policies}
%
%\bigskip\noindent
%Nature Portfolio journals: \url{https://www.nature.com/nature-research/editorial-policies}
%
%\bigskip\noindent
%\textit{Scientific Reports}: \url{https://www.nature.com/srep/journal-policies/editorial-policies}
%
%\bigskip\noindent
%BMC journals: \url{https://www.biomedcentral.com/getpublished/editorial-policies}
%\end{flushleft}

\begin{appendices}

%\section{Section title of first appendix}\label{secA1}

%An appendix contains supplementary information that is not an essential part of the text itself but which may be helpful in providing a more comprehensive understanding of the research problem or it is information that is too cumbersome to be included in the body of the paper.
% \newpage
% \clearpage
\section*{Appendix}
\label{sec:appendix}
\appendix
\input{sections/supplemental_material}




\end{appendices}


\bibliographystyle{bst/sn-mathphys}
\bibliography{bib/main}

%\bibliography{sn-bibliography}


\end{document}

%% file: sections/introduction.tex
\section{Introduction}
\label{sec:introduction}

Natural disasters cause significant damage (e.g., Hurricane Harvey in 2017 cost \$125 billion)\footnote{\url{https://en.wikipedia.org/wiki/List_of_disasters_by_cost}} and require urgent assistance in time of crisis. In the last decade, various social media played important roles in humanitarian response tasks as they were widely used to disseminate information and obtain valuable insights. During disaster events, people post content (e.g., text, images, and video) on social media to ask for help (e.g., report of a person stuck on a rooftop during a flood), offer support, identify urgent needs, or share their feelings. Such information is helpful for humanitarian organizations to take immediate actions to plan and launch relief operations.
Recent studies demonstrated that images shared on social media during a disaster can assist humanitarian organizations in recognizing damages in infrastructure~\cite{Mouzannar2018}, assessing damage severity~\cite{nguyen17damage}, identifying humanitarian information~\cite{alam2018crisismmd}, detecting crisis incidents~\cite{weber2020detecting}, and detecting disaster events with other related tasks~\cite{FAlam:ASONAM20}. However, the amount of research and resources to develop powerful computer vision-based predictive models remains insufficient compared to the NLP-based progress~\cite{imran2015processing,said2019natural,Imran:IPM20}. Motivated by these observations, this research aims to enrich available resources to make further advancements in the computer vision-based disaster management studies.
%

% Several task-specific models need to be deployed to track real-time disaster events and extract humanitarian and damage-related information~\cite{alam2018Image,alam17demo}. These tasks include {\em (i)} disaster type recognition, {\em (ii)} informativeness classification, {\em (iii)} humanitarian categorization, and {\em (iv)} damage severity assessment (see Section~\ref{sec:tasks_dataset} for more details). Existing works \cite{nguyen17damage,alam2018crisismmd,Mouzannar2018} address these tasks separately, resulting in higher computational complexities (e.g., computational power, training and inference time). Hence, this research aims at reducing this overhead by addressing different tasks simultaneously in a multi-task learning (MTL) setup, which can also help reduce the carbon footprint~\cite{schwartz2020green}.

Recent advances in deep convolutional neural networks (CNN) and their learning techniques provide efficient solutions for different computer vision applications. While simple applications can be realized with a single-task formulation such as classification~\cite{he2016deep}, semantic segmentation~\cite{long2015fully}, or object detection~\cite{redmon2016you}, the complex ones such as autonomous vehicles, robotics, and social media image analysis~\cite{alam2018Image,yu2020bdd100k} necessitate incorporating multiple tasks, which significantly increases the computational and memory requirements for both training and inference. Multi-task learning (MTL) techniques \cite{caruana1997multitask, yu2020bdd100k, vandenhende2021multi} have emerged as the standard approach for these complex applications where a model is trained to solve multiple tasks simultaneously, which helps to improve the performance, reduce inference time and computational complexities. For example, an image posted on social media during a disaster event may contain information whether it is a flood event, shows infrastructure damage, and is severe. Such a multitude of information needs to be detected in real-time to help humanitarian organizations \cite{alam2018Image,alam17demo} with various tasks 
% where a single model solving multiple tasks can be more effective than having multiple models for multiple tasks. 
%
% Typically several task-specific models need to be deployed to track real-time disaster events and extract humanitarian and damage-related information~\cite{alam2018Image,alam17demo}. These tasks 
including {\em (i)} disaster type recognition, {\em (ii)} informativeness classification, {\em (iii)} humanitarian categorization, and {\em (iv)} damage severity assessment (see Section~\ref{sec:tasks_dataset} for more details). Existing works \cite{nguyen17damage,alam2018crisismmd,Mouzannar2018} present separate task-specific models, resulting in higher computational complexities (e.g., computational power, training and inference time). Hence, this research aims at reducing this overhead by addressing different tasks simultaneously with an MTL setup, which can also help reduce the carbon footprint~\cite{schwartz2020green}.

%% Necessity of dataset construction
Labeled public image datasets, such as ImageNet \cite{ILSVRC15} and Microsoft COCO \cite{coco_2014} made significant contributions to the advancement of today’s powerful machine learning models. Likewise, for the MTL setup, several image datasets have already been proposed, which are summarized in Table \ref{tab:available_datasets}. These datasets include images from different domains such as indoor scenes, driving, faces, handwritten digits, and animal recognition, which are already contributing to the advancement of MTL research. However, an MTL dataset for critical real-world applications which comprise humanitarian response tasks during natural disasters is yet to become available. This paper proposes a novel MTL dataset for disaster image classification.

\begin{figure}[t]
\centering
\includegraphics[width=0.9\textwidth]{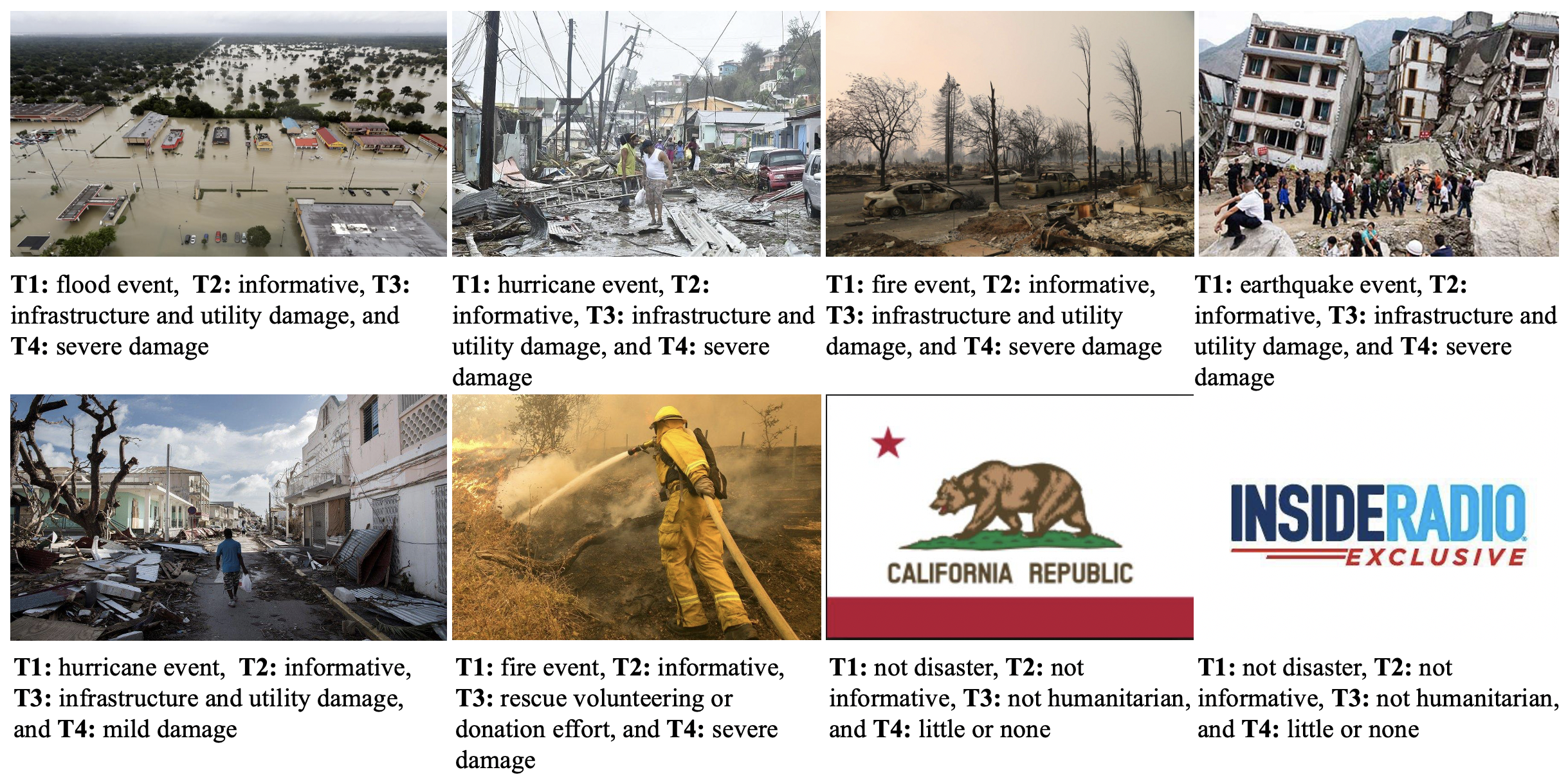}
\caption{Examples of images representing all tasks. \textbf{T1}: Disaster types,  \textbf{T2}: Informativeness, \textbf{T3}: Humanitarian, \textbf{T4}: Damage severity.}
\label{fig:example_all_task}
\end{figure}

To this end, we build upon the previous work of Alam et al. \cite{FAlam:ASONAM20} where the images are mostly annotated for individual tasks, and only 5,558 out of 71,198 images have labels for all four tasks mentioned above. We provide an expansive extension by annotating the images for all tasks, i.e., we annotated 155,899 more labels for these tasks in addition to the existing ones.\footnote{For four tasks, 71,198 images now have 284,792 labels whereas previous annotations comprised only 128,893 labels.} \rev{For \textit{disaster type recognition} and \textit{humanitarian categorization} tasks, we also labeled a part of the images with multiple labels following a weak supervision approach as they are suitable for multilabel annotation (see Section~\ref{sec:tasks_dataset}).} Figure \ref{fig:example_all_task} shows example images with the labels for all four tasks.

Our contributions in this research can be summarized as follows: {\em(i)} we provide a social media MTL image dataset for disaster response tasks with various complexities, which can be used as an evaluation benchmark for computer vision research; {\em(ii)} we ensured high quality annotations by making sure that at least two annotators agree on a label; {\em (iii)} we provide a benchmark for heterogeneous multi-task learning and baseline studies to facilitate future study; {\em(iv)} our experimental results can also be used as a baseline in the single-task learning setting. 
%%%%%%% Important limitation

% %%%%%%%paper structure
The rest of the paper is organized as follows. Section~\ref{sec:related_works} provides an overview of the existing work. Section~\ref{sec:tasks_dataset} introduces the tasks and describes the dataset development process. Section~\ref{sec:experiments} explains the experiments and presents the results while Section~\ref{sec:discussions} provides a discussion. Finally, we conclude the paper in Section~\ref{sec:conclutions}.

%% file: sections/related_work.tex
\section{Related Work}
\label{sec:related_works}
This paper mainly focuses on the development of an MTL dataset for disaster response tasks. Therefore, we first review the recent work on MTL and available MTL datasets; and then, survey social media image classification literature and datasets for disaster response.

\subsection{Multi-Task Learning and Datasets}
Multi-task learning (MTL) aims to improve generalization capability by leveraging information in the training data consisting of multiple related tasks \cite{caruana1997multitask}. It simultaneously learns multiple tasks and has shown promising results in terms of generalization, computation, memory footprint, performance, and inference time by jointly learning through a shared representation \cite{caruana1997multitask,vandenhende2021multi}. Since the seminal work by Caruana \cite{caruana1997multitask}, MTL research has received wide attention in the last several years in NLP, computer vision, and other research areas~\cite{ruder2017overview,zhang2021survey,vandenhende2021multi,crawshaw2020multi,worsham2020multi}. MTL brings benefits when associated tasks share complementary information. However, performance can suffer when multiple tasks have conflicting needs, and the tasks have competing priorities (i.e., one is superior to the other). This phenomenon is referred to as negative transfer. This understanding led to the question of what, when, and how to share information among tasks \cite{strezoski2019learning,vandenhende2021multi}. To address these aspects, in the deep learning era, numerous architectures and optimization methods have been proposed. The architectures are categorized into hard and soft parameter sharing. Hard parameter sharing design consists of a shared network followed by task-specific heads \cite{kokkinos2017ubernet,kendall2018multi,chen2018gradnorm}. In soft parameter sharing, each task has its own set of parameters, and a feature sharing mechanism to deal with cross-task talk \cite{misra2016cross,ruder2019latent,gao2019nddr}. 
In MTL literature, a problem can be formulated in two different ways - homogeneous and heterogeneous \cite{strezoski2019learning}. While the homogeneous MTL assumes that each task corresponds to a single output, the heterogeneous MTL assumes each task corresponds to a unique set of output labels \cite{caruana1997multitask,yang2017deep}. The latter setting uses a neural network using multiple sets of outputs and losses. 
In this study, we aim to provide a benchmark with our heterogeneous MTL dataset using the hard parameter sharing approach.

% In the multi-task learning literature, a problem can be formulated in two different ways - homogeneous and heterogeneous \cite{strezoski2019learning}. While the homogeneous multi-task learning assumes that each task corresponds to a single output, the heterogeneous multi-task learning assumes each task corresponds to a unique set of output labels \cite{caruana1997multitask,yang2017deep}. The latter setting uses a neural network using multiple sets of outputs and losses. 
% The multi-task learning approaches are categorized in hard and soft parameter sharing \cite{caruana1997multitask}. Hard parameter sharing is the commonly used neural networks based approach, which is applied by sharing the hidden layers between all tasks and keeping the several task-specific output layers \cite{caruana1997multitask}. This research formulates the problem based on the heterogeneous learning setting and address it using the hard-parameter sharing based neural network setting. 
%We provide benchmarks results and present single and multitask learning results using several pre-trained network architectures as backbone models.

% \paragraph{Datasets}
% Bdd100k: A diverse driving dataset for heterogeneous multitask learning
% \cite{yu2020bdd100k}
Earlier studies such as \cite{kang2011learning} and \cite{kumar2012learning} mostly exploited the MNIST \cite{lecun1998gradient} and USPS \cite{hull1994database} datasets for MTL experiments. These datasets were originally designed for single-task classification settings. For example, the widely used MNIST dataset was originally designed for digit classification, and Office-Caltech \cite{gong2012geodesic} was designed to categorize images in 31 classes, which are collected from different domains. However, such datasets are used with the homogeneous problem setting of multi-task learning by selecting 10 target classes as 10 binary classification tasks \cite{kumar2012learning,strezoski2019learning,yang2016deep}. Numerous other widely used datasets such as MC-COCO \cite{coco_2014} and CelebA \cite{liu2015deep} have also been used for multi-task learning in the homogeneous problem setting. 

Several existing datasets consisting of multiple unique output label sets were studied in the heterogeneous setting. For example, AdienceFaces \cite{eidinger2014age} was designed for gender and age group classification tasks, OmniArt \cite{strezoski2018omniart} consists of seven tasks, NYU-V2 \cite{silberman2012indoor} consists of three tasks, and PASCAL \cite{everingham2010pascal,chen2014detect} consists of five tasks. Very few datasets were specifically designed for multi-task learning research. Most notable ones are Taskonomy \cite{zamir2018taskonomy} and BDD100K \cite{yu2020bdd100k}. The Taskonomy dataset consists of four million images of indoor scenes from 600 buildings, and each image was annotated for twenty-six visual tasks. Ground truths of this dataset were obtained programmatically, and knowledge distillation approaches. The BDD100K dataset is a diverse 100K driving video dataset consisting of ten tasks. It was collected from Nexar,\footnote{\url{https://www.getnexar.com/}} where videos are uploaded by the drivers. In Table \ref{tab:available_datasets}, we provide widely used datasets, which have been used for MTL.

\begin{table}[htb!]
\centering
\scalebox{0.66}{
% \resizebox{\textwidth}{!}{%
\setlength{\tabcolsep}{2.0pt}
\begin{tabular}{@{}llllllllll@{}}
\toprule
\multicolumn{1}{@{}l}{\textbf{Ref.}} & \multicolumn{1}{l}{\textbf{Dataset}} & \multicolumn{1}{l}{\textbf{Source}} & \multicolumn{1}{l}{\textbf{Size}} & \multicolumn{1}{l}{\textbf{Task type}} & \multicolumn{1}{l}{\textbf{\# Tasks}} & \multicolumn{1}{l}{\textbf{Tasks}} & \multicolumn{1}{l}{\textbf{\# Classes}} & \multicolumn{1}{l}{\textbf{Domain}} & \multicolumn{1}{l@{}}{\textbf{Year}} \\ \midrule
\multicolumn{10}{@{}l}{\textbf{Datasets used for multi-task learning}}  \\ \midrule
\cite{vandenhende2021multi} & PASCAL \cite{everingham2010pascal,chen2014detect} & Flickr & 12,030 (I) & Hete. & 5 & \begin{tabular}[c]{@{}l@{}} SS, HS, \\SE, and SD\end{tabular} & - & \begin{tabular}[c]{@{}l@{}}Diverse\\ objects\end{tabular} & 2021 \\
\cite{vandenhende2021multi} & NYU-V2 \cite{silberman2012indoor} & PC & 1,449 (I) & Hete. & 3 & \begin{tabular}[c]{@{}l@{}}IS, SS, \\and SC\end{tabular} & - & Indoor video & 2021 \\
\cite{yu2020bdd100k} & BDD100K & Nexar & 100,000 (V) & Hete. & 10 & ten tasks & \footnote{https://doc.bdd100k.com/format.html\#categories} & Driving & 2020 \\
\cite{strezoski2019learning} & MNIST \cite{lecun1998gradient} & - & 70,000 (I) & Homo. & 10 & 10 digits cls. & 10 CL. & Handwritten & 2019 \\
\cite{strezoski2019learning} & CIFAR10 \cite{krizhevsky2009learning} & - & 60,000 (I) & Homo. & 10 & 10 animal cls. & 10 CL. & Animal & 2019 \\
\cite{strezoski2019learning} & UCSD-Birds \cite{wah2011caltech} & - & 11,788 (I) & Homo. & 10 & 10 R/tasks & Ranking & Animal & 2019 \\
\cite{strezoski2019learning} & OmniGlot \cite{lake2015human} & - & 1,623 (I) & Homo. & 50 & 50 alphabets & 50 CL. & Handwritten & 2019 \\
\cite{strezoski2019learning} & OmniArt \cite{strezoski2018omniart} &  & 133,000 (S) & Hete. & 7 & 7 tasks & - & Artwork & 2019 \\
\cite{zamir2018taskonomy} & Taskonomy & IC & 4M (I) & Hete. & 26 & 26 tasks & - & Indoor scenes & 2018 \\
\cite{long2017learning} & Office-caltech \cite{gong2012geodesic} &  & 2,533 (I) & Homo. & 4 & \begin{tabular}[c]{@{}l@{}}Amazon, Webcam,\\ and DSLR, \\Caltech-256\end{tabular} & 10 CL/task &  & 2017 \\
\cite{long2017learning} & Office-Home \cite{venkateswara2017deep} & SE & 15,500 (I) & Homo. & 4 & \begin{tabular}[c]{@{}l@{}}Artistic, clip art,\\ product, and\\ real-world images\end{tabular} & 65 objects & Office/Home & 2017 \\
\cite{long2017learning} & ImageCLEF\footnote{http://imageclef.org/2014/adaptation} & - & 2,400 (I) & Homo. & 4 & \begin{tabular}[c]{@{}l@{}}Caltech-256, \\ImageNet\\ Pascal and Bing\end{tabular} & - & Diverse & 2017 \\
\cite{yang2016deep} & MNIST \cite{lecun1998gradient} & - & 70,000 (I) & Homo. & 10 & 10 digits cls. & 10 CL. & Handwritten & 2016 \\
\cite{yang2016deep} & AdienceFaces \cite{eidinger2014age} & Flickr & \begin{tabular}[c]{@{}l@{}}16,252 (I) (G),\\ 16,139 (I) (A)\end{tabular} & Hete. & 2 & Gender, Age & \begin{tabular}[c]{@{}l@{}}Gender: 2\\ Age: 8\end{tabular} & Face & 2016 \\
\cite{kumar2012learning} & USPS \cite{hull1994database} & - & 2,000 (S) & Homo. & 10 & 10 ways/tasks & digits: 0-9 & Handwritten & 2012 \\
\cite{kumar2012learning} & MNIST \cite{lecun1998gradient} & - & 2,000 (S) & Homo. & 10 & 10 ways/tasks & digits: 0-9 & Handwritten & 2012 \\
\cite{kang2011learning} & USPS \cite{hull1994database} & - & 2,000 (S) & Homo. & 10 & 10 ways/tasks & digits: 0-9 & Handwritten & 2011 \\
\cite{kang2011learning} & MNIST \cite{lecun1998gradient} & - & 2,000 (S) & Homo. & 10 & 10 ways/tasks & digits: 0-9 & Handwritten & 2011 \\
\cite{kang2011learning} & Animal \cite{lampert2009learning} & - & 30,000 (I) & Homo. & 20 & 20 ways/tasks & 20 CL. & Animal & 2011 \\ 
\midrule
\multicolumn{10}{@{}l}{\textbf{Disaster-related datasets}} \\ \midrule
\cite{weber2020detecting} & Incident & Web, SM & 446,684 (I) & NA & 1 & Incident & 43 & Incidents & 2020 \\
\cite{FAlam:ASONAM20}  & CrisisBench. & Web, SM & \begin{tabular}[c]{@{}l@{}}DT:17,511, \\Info:59,717,\\ Hum:17,769, \\DS:34,896 \end{tabular}& NA & 4 & \begin{tabular}[c]{@{}l@{}}DT, Info, \\Hum, DS\end{tabular} & \begin{tabular}[c]{@{}l@{}}DT: 7, Info: 2,\\ Hum:4,\\ DS:3\end{tabular} & Disaster & 2020 \\
% \tablefootnote{5,558 images for 4 tasks.}
% \tablefootnote{3,533 images for 3 tasks.}
\cite{Gupta_2019_CVPR_Workshops} & xBD & Satellite & 700,000 & NA & - & Building damage & 4 & Disaster & 2019\\
\cite{alex258247} & MediaEval 2018 & SM & 1,654 I/P & NA & 1 & Flood & R. and cls.: 2 CL. & Disaster & 2018 \\
\cite{alam2018crisismmd} & CrisisMMD & SM & 18,082 & NA & 3 & Info, Hum, DS &\begin{tabular}[c]{@{}l@{}} Info: 2, Hum:8,\\ DS:3 \end{tabular}& Disaster & 2018 \\
\cite{Mouzannar2018} & DMD & Web & 5878 & NA & 1 & Damage & 6 & Disaster & 2018 \\
\cite{nguyen2017automatic} & DAD & SM & $\sim$25,000 & NA & 1 & DS & 3 & Disaster & 2017 \\
\cite{bischke2017multimedia} & DIRSM & Flickr & \begin{tabular}[c]{@{}l@{}}T1: 6,600 (I);\\ T2: 462 I/P \end{tabular} & NA & 1 & Flood & R, cls.: 2 CL & Disaster & 2017 \\  \midrule
\multicolumn{10}{@{}l}{\textbf{Proposed disaster-related multi-task learning dataset}} \\ \midrule
 & MEDIC & SM & 71,198 (I) & Hete. & 4 & DT, Info, Hum, DS & \begin{tabular}[c]{@{}l@{}}DT: 7, Info: 2,\\ Hum:4, DS:3\end{tabular} & Disaster & 2021 \\ \bottomrule
\end{tabular}%
}
\caption{Upper part of the table presents the datasets used in multi-task learning studies in computer vision research. Middle part shows disaster related datasets, and the last row shows our proposed dataset. I: Images, V: Videos, S: Samples, SE: Search engines, SM: Social media, DT: disaster types, Info: Informativeness, Hum: Humanitarian, DS: Damage severity. CL.: number of class labels. Hete.: heterogeneous, Homo: Homogeneous. PC: Personal collection. SS: semantic segmentation, HS: human part segmentation, SE: semantic edge detection of surface normals prediction, SD: saliency detection, IS: instance segmentation, SC: scene classification, IC: Indoor scenes, cls.: classification, R/tasks: Ranking tasks, I/P: image patches.}
\label{tab:available_datasets}
% \vspace{-2em}
\end{table}

\subsection{Disaster Response Studies and Datasets}
\label{ssec:disaster_response_studies}
% \paragraph{Social Media Images for Disaster Response}
% \paragraph{Social Media Content}
% \rev{
During disaster events, social media content has proven to be effective in facilitating different stakeholders including humanitarian organizations \cite{Imran2022}. Alongside, there has been growing research interest in developing computational methods and systems to better analyze and extract actionable information from social media content \cite{alam-etal-2018-domain,said2019natural,Alam2019}. Most of such efforts relied on social media content, such as Twitter and Facebook, for humanitarian aid ~\cite{imran2014aidr,Villegas2018}. Given that accessing Facebook data became difficult, the use of Twitter content remained more popular. Research studies and resource development have focused on Twitter content due to its instant access to timely multi-modal information (i.e., textual and visual) as such information is crucial for different stakeholders (e.g., governmental and non-governmental organizations)~\cite{imran2014aidr,Villegas2018}. Notable resources with textual content include the CrisisLex~\cite{olteanu2014crisislex}, CrisisNLP \cite{imran2016lrec}, TREC Incident Streams~\cite{mccreadie2019trec}, disaster tweet corpus~\cite{Wiegmann2020}, Arabic Tweet Corpus~\cite{alharbi2019crisis}, CrisisBench~\cite{alam2020standardizing}, HumAID~\cite{alam2021humaid}, and CrisisMMD (text and image)\cite{alam2018crisismmd,ofli2020analysis}.
% }
% \rev{
In the past years, several systems have also been developed and deployed during disaster events \cite{imran2014aidr,burel2017semantics,burel2018crisis,imran2020rapid}. One notable system is AIDR~\cite{imran2014aidr}\footnote{\url{http://aidr.qcri.org/}}, which has been used during major disaster events to collect and classify tweets, and provide a visual summary.
% }

% \rev{
Earlier research efforts in crisis informatics are mainly focused on textual content analysis \cite{Imran:IPM20}. However,
lately there has been a growing interest on the imagery content analysis as
% \cite{nguyen17damage,Imran:IPM20,Mouzannar2018,FAlam:ASONAM20,weber2020detecting}. The key applications for disaster-related social media imagery include severity assessment and identification of damaged \cite{nguyen17damage,Mouzannar2018}, filtering humanitarian information \cite{alam2018crisismmd}, crisis incidents detection \cite{weber2020detecting}, and detecting disaster events with other related tasks~\cite{FAlam:ASONAM20}. 
images posted on social media during disasters can play significant role as reported in many studies~\cite{petersinvestigating,daly2016mining,nguyen2017automatic,nguyen17damage,alam17demo,alam2018Image}. Recent works include categorizing the severity of damage into discrete levels~\cite{nguyen2017automatic,nguyen17damage,alam17demo} or quantifying the damage severity as a continuous-valued index~\cite{nia2017building,li2018localizing}. Such models were also used in real-time disaster response scenarios by engaging with emergency responders \cite{imran2020rapid}. Other related work includes adversarial networks for data scarcity issues~\cite{li2019identifying,pouyanfar2019unconstrained}; disaster image retrieval~\cite{ahmad2017convolutional}; image classification in the context of bush fire emergency~\cite{10.3389/frobt.2016.00054}; flood photo screening system \cite{ning2020prototyping}; sentiment analysis from disaster image \cite{hassan2019sentiment}; monitoring natural disasters using satellite images \cite{ahmad2017jord,said2019natural}; and flood detection using visual features \cite{Jony2019FloodDI}.
% }

% \paragraph{Image Datasets}
% In crisis informatics\footnote{\url{https://en.wikipedia.org/wiki/Disaster_informatics}} research the 
Publicly available image datasets include damage severity assessment dataset (DAD)~\cite{nguyen17damage}, multimodal dataset (CrisisMMD) ~\cite{alam2018crisismmd} and damage identification multimodal dataset (DMD)~\cite{Mouzannar2018}. The first dataset is only annotated for images, whereas the last two are annotated for both text and images. Other relevant datasets are Disaster Image Retrieval from Social Media (DIRSM) \cite{bischke2017multimedia} and MediaEval 2018 \cite{alex258247}. The dataset reported in \cite{Gupta_2019_CVPR_Workshops} was constructed for detecting damage as an anomaly using pre-and post-disaster images. It consists of 700,000 building annotations. A similar and relevant work is the Incidents dataset \cite{weber2020detecting}, which consists of 446,684 manually labeled images with 43 incident categories. The \textit{Crisis Benchmark Dataset} reported in \cite{FAlam:ASONAM20} is the largest social media disaster image classification dataset, which is a consolidated version of DAD, CrisisMMD, DMD, and additional labeled images. 

In this study, we extended the \textit{Crisis Benchmark Dataset} to adapt it to an MTL setup. To that end, we assigned images with 155,899 more labels to ensure that the entire dataset contains aligned labels for all the tasks. 
% \rev{
Additionally, we annotated some images with multiple labels, when appropriate, for humanitarian categorization and disaster type recognition tasks.
% }

%% file: sections/dataset.tex
\section{MEDIC Dataset}
\label{sec:tasks_dataset}

The MEDIC dataset consists of four different disaster-related tasks that are important for humanitarian aid.\footnote{\url{https://en.wikipedia.org/wiki/Humanitarian_aid}} These tasks are defined based on prior work experience with the humanitarian response organizations such as UN-OCHA and existing literature \cite{imran2014aidr,imran2015processing,alam2018crisismmd,alam2018Image}. In this section, we first provide the details of each task and class labels, and then, discuss the annotation details of the dataset.

\subsection{Tasks}
\label{ssec:tasks}
\paragraph{Disaster Types}
\label{sssec:task_disaster_types}
During man-made and natural disasters, people post textual and visual content about the current situation, and the real-time social media monitoring system requires to detect an event when ingesting images from unfiltered social media streams. For the disaster scenario, it is important to automatically recognize different disaster types from the crawled social media images. For instance, an image can depict a wildfire, flood, earthquake, hurricane, and other types of disasters. Different categories (i.e., natural, human-induced, and hybrid) and sub-categories of disaster types have been defined in the literature \cite{shaluf2007disaster}. 
% Three major categories include natural, human-induced, and hybrid. Each of them has also been classified into sub-categories. 
This research focuses on major disaster events that include {\em (i)} earthquake, {\em (ii)} fire, {\em (iii)} flood, {\em (iv)} hurricane, {\em (v)} landslide, {\em (vi)} other disaster, which covers all other types (e.g., plane, train crash), and {\em (vii)}  not disaster, which includes the images that do not show any identifiable disasters.

\paragraph{Informativeness}
\label{sssec:task_informativeness}
Social media contents are often noisy and contain numerous irrelevant images such as cartoons, advertisements, etc. In addition to this, the clean images that show damaged infrastructure due to flood, fire, or any other disaster events are crucial for humanitarian response tasks. Therefore, it is necessary to eliminate any irrelevant or redundant content to facilitate crisis responders' efforts more effectively. For this purpose, we define the \textit{informativeness} task as to filter out irrelevant images, where the class labels comprise {\em (i)} informative and {\em (ii)} not informative.

\paragraph{Humanitarian}
\label{sssec:task_humanitarian}
Fine-grained categorization of certain information significantly helps the emergency crisis responders to make an efficient actionable decision.
Humanitarian categories vary depending on the type of content (text vs. image). For example, the CrisisBench dataset \cite{alam2020standardizing} consists of tweets labeled with 11 categories, whereas CrisisMMD \cite{alam2018crisismmd} multimodal dataset consists of 8 categories. Such variation exists between text and images because some information can easily be presented in one modality than another modality. For example, it is possible to report \textit{missing or found people} in text than in an image, which is also reported in \cite{alam2018crisismmd}. This research focuses on these factors and considers the four most important categories that are useful for crisis responders such as {\em (i)} affected, injured, or dead people, {\em (ii)} infrastructure and utility damage, {\em (iii)} rescue volunteering or donation effort, and {\em (iv)} not humanitarian.

\paragraph{Damage Severity}
\label{ssec:task_damage_severity}
Detecting the severity of the damage is significantly important to help the affected community during disaster events. The severity of the damage can be assessed from an image based on the visual appearance of the physical destruction of a built structure (e.g., bridges, roads, buildings, burned houses, and forests). In line with~\cite{nguyen17damage}, this research defines the following categories for the classification task: {\em (i)} severe damage, {\em (ii)} mild damage, and {\em (iii)} little or none.

\subsection{Annotations}
\label{ssec:dataset}

\subsubsection{Data Curation}
This research extends the labels of the Crisis Benchmark dataset~\cite{FAlam:ASONAM20}. The Crisis Benchmark dataset was developed by consolidating existing datasets and labeling new data for disaster types. 
% The existing datasets include DAD~\cite{nguyen17damage}, CrisisMMD~\cite{alam2018crisismmd} and DMD~\cite{Mouzannar2018} datasets. The DMD dataset has relabeled for all four tasks. New data has also been annotated for disaster type and informativeness tasks. Such detail can be found in \cite{FAlam:ASONAM20}.  
% For the sake of clarity and completeness, we provide a brief overview of the dataset. 
The Crisis Benchmark dataset consists of images collected from Twitter, Google, Bing, Flickr, and Instagram. The majority of the datasets have been collected from Twitter, as shown in Table \ref{tab:dataset_detail}. The Twitter data were mainly collected during major disaster events\footnote{Event names reported in Table \ref{tab:dataset_detail} are based on Wikipedia.} and using different disaster-specific keywords. The data collected from Google, Bing, Flickr, and Instagram are based on specific keywords. 
% For clarity we provided such keywords lists in Appendix (see section \ref{appendix:data_collection}). 
The dataset is diverse in terms of {\em(i)} number of events, {\em(ii)} different time frames spanning over five years, {\em(iii)} natural (e.g., earthquake, fire, floods) and man-made disasters (e.g., Paris attack, Syria attacks), and {\em(iv)} events occurred in different parts of the world. The number of images in different events resulted from different factors, such as the number of tweets collected during the disaster events, the number of images crawled, filtered due to duplicates, and a random selection for the annotation. Our motivation for choosing and extending the Crisis Benchmark dataset is that it reduced the overall cost of data collection and annotation processes while also having a large dataset for MTL.

\begin{table}[]
\centering
\scalebox{0.76}{
\setlength{\tabcolsep}{2.5pt}
\begin{tabular}{@{}llrr|llrr@{}}
\toprule
\multicolumn{1}{@{}l}{\textbf{Source}} & \multicolumn{1}{l}{\textbf{Event name}} & \multicolumn{1}{r}{\textbf{Year}} & \multicolumn{1}{r|}{\textbf{\# Images}} & \multicolumn{1}{l}{\textbf{Source}} & \multicolumn{1}{l}{\textbf{Event name}} & \multicolumn{1}{r}{\textbf{Year}} & \multicolumn{1}{r@{}}{\textbf{\# Images}} \\ \midrule
Twitter & Typhoon ruby/hagupit & 2014 & 833 & Twitter & Iraq iran earthquake & 2017 & 596 \\
Twitter & Nepal earthquake & 2015 & 21,710 & Twitter & Mexico earthquake & 2017 & 1,378 \\
Twitter & South India floods & 2015 & 1,476 & Twitter & Srilanka floods & 2017 & 1,022 \\
Twitter & Illapel earthquake & 2015 & 403 & Twitter & Ukraine conflict & 2017 & 240 \\
Twitter & Food insecurity in yemen & 2015 & 466 & Twitter & Greece wildfire & 2018 & 351 \\
Twitter & Paris attack & 2015 & 1,043 & Twitter & Hurricane florence & 2018 & 186 \\
Twitter & South India floods & 2015 & 753 & Twitter & Hurricane michael & 2018 & 219 \\
Twitter & Syria attacks & 2015 & 350 & Twitter & Kerala flood & 2018 & 605 \\
Twitter & Terremotoitalia & 2015 & 919 & Twitter & Typhoon mangkhut & 2018 & 172 \\
Twitter & Ecuador earthquake & 2016 & 2,280 & Google & NA & NA & 3,007 \\
Twitter & Hurricane matthew & 2016 & 596 & Twitter & Human induced disaster & NA & 501 \\
Twitter & California wildfires & 2017 & 1,585 & \begin{tabular}[c]{@{}l@{}}G, B, F\end{tabular}  & NA & NA & 1,263 \\
Twitter & Hurricane harvey & 2017 & 5,644 & Twitter & Natural disaster & NA & 6,597 \\
Twitter & Hurricane irma & 2017 & 4,973 & Twitter & Security incidents activities & NA & 1,082 \\
Twitter & Hurricane maria & 2017 & 5,069 & G, I. & NA & NA & 5,879 \\ \bottomrule
\end{tabular}%
}
\caption{Data collection source, event name, year of the event and number of image annotated. G: Google, B: Bing, F: Flickr, I: Instagram.}
\label{tab:dataset_detail}
\vspace{-2em}
\end{table}

% \subsubsection{Annotation}
\subsubsection{Multiclass Annotation}
\label{ssec:annotation}
For the manual annotation, we used Appen\footnote{\url{https://appen.com/}} crowdsourcing annotation platform. In such a platform, finding qualified workers and managing the quality of the annotation is an important issue. To ensure the quality, we used the widely used gold standard evaluation approach \cite{chowdhury2015selection}. We designed the interface with annotation guidelines on Appen for the annotation task (see Figure \ref{fig:example_annotation_interface} in Appendix). We followed the annotation guidelines from previous work \cite{alam2018crisismmd,FAlam:ASONAM20} and improved with examples for this task (see the detailed annotation guidelines with examples in Appendix \ref{appendix:data_collection}). 

% For all tasks, we choose to annotate in a multiclass setting even though \textit{humanitarian} and \textit{disaster type} tasks in our context are more suitable to be framed as pure multi-label.
% rather than multi-class problems. 
% Our decision has been influenced by several factors. The most important one was our consultation with humanitarian organizations which suggested limiting the number of classes by merging related ones and keeping only the most important information types. This is due to the information overload issue that humanitarian responders often deal with at the onset of a disaster situation if exposed to information types not important for them. Furthermore, obtaining a sufficient number of labeled instances for a large number of classes to train a pure multi-label classifier is not practical due to both annotation budget (e.g., time, cost) and modeling perspectives (e.g., high imbalance). For the image, which can have multiple labels, we instructed the annotators to select the label that is more important for humanitarian organizations and prominent in the image. 

\rev{
For all tasks, we first annotated images with a multiclass setting. Then for \textit{humanitarian} and \textit{disaster type} tasks we labeled the images with multiple labels as they are more suitable to be framed as pure multilabel setting (see Section \ref{sssec:multilabel_annotation}). 
}
\rev{
% \textbf{Multiclass Annotation:} 
For the multiclass labeling, our decision has been influenced by several factors. The most important one was our consultation with humanitarian organizations which suggested limiting the number of classes by merging related ones and keeping only the most important information types. This is due to the information overload issue that humanitarian responders often deal with at the onset of a disaster situation if exposed to information types not important for them. For an image that can have multiple labels, we instructed the annotators to select the label that is more important for humanitarian organizations and prominent in the image.
}

For the annotation, we designed a {\em HIT} containing five images. For the gold standard evaluation, we manually labeled 100 images, which are randomly assigned to the HIT for the evaluation. We assigned a criterion to have at least three annotations per image and per task. An agreement score of 66\% is used to select the final label, which ensured that at least two annotators agreed on a label. The HIT was extended to more annotators if such a criterion was not met. 

Since the Crisis Benchmark dataset did have task-specific labels for all images, i.e., different sets of images consisted of labels for three tasks and two tasks, we first prepared different sets with missing labels for the annotation. For example, 25,731 images of the Crisis Benchmark dataset did not have labels for disaster types and humanitarian tasks, which we selected for the annotation tasks. In this way, we run the annotation tasks in different batches.

\begin{table}[]
\centering
% \resizebox{\textwidth}{!}{%
\scalebox{0.85}{
\setlength{\tabcolsep}{2pt}
\begin{tabular}{@{}lrrr|lrrr@{}}
\toprule
\multicolumn{1}{@{}l}{\textbf{Tasks}} & \multicolumn{1}{r}{\textbf{Fleiss ($\kappa$)}} & \multicolumn{1}{r}{\textbf{Krip. ($\alpha$)}} & \multicolumn{1}{r|}{\textbf{Avg agg.}} & \multicolumn{1}{l}{\textbf{Tasks}} & \multicolumn{1}{r}{\textbf{Fleiss ($\kappa$)}} & \multicolumn{1}{r}{\textbf{Krip. ($\alpha$)}} & \multicolumn{1}{r@{}}{\textbf{Avg agg.}} \\ \midrule
Disaster types & 0.46 & 0.46 & 0.70 & Humanitarian & 0.52 & 0.52 & 0.73 \\
Informativeness & 0.71 & 0.71 & 0.91 & Damage severity & 0.55 & 0.55 & 0.79 \\ \bottomrule
\end{tabular}%
}
\caption{Annotation agreement for different tasks. Fleiss Kappa ($\kappa$), Krip. ($\alpha$): Krippendorff's $\alpha$, Avg agg.: Average observed agreement.}
\label{tab:annotation_agreement}
% \vspace{-0.5em}
\end{table}

% \usepackage{booktabs}
\begin{table}[]
\centering
\scalebox{0.8}{
\setlength{\tabcolsep}{2.0pt}
\begin{tabular}{@{}lrrrr|lrrrr@{}}
\toprule
\multicolumn{1}{@{}l}{\textbf{Label}} & \multicolumn{1}{r}{\textbf{Train}} & \multicolumn{1}{r}{\textbf{Dev}} & \multicolumn{1}{r}{\textbf{Test}} & \multicolumn{1}{r|}{\textbf{Total}} & \multicolumn{1}{l}{\textbf{Label}} & \multicolumn{1}{r}{\textbf{Train}} & \multicolumn{1}{r}{\textbf{Dev}} & \multicolumn{1}{r}{\textbf{Test}} & \multicolumn{1}{r@{}}{\textbf{Total}} \\ \midrule
\multicolumn{5}{c|}{\cellcolor[HTML]{DAE8FC}\textbf{Disaster Types}} & \multicolumn{5}{c}{\cellcolor[HTML]{DAE8FC}\textbf{Humanitarian}} \\ \midrule
Earthquake & 12,296 & 1,004 & 1,795 & 15,095 & Affected injured or dead people & 3,662 & 274 & 639 & 4,575 \\
Fire & 1,796 & 262 & 690 & 2,748 & Infrastructure and utility damage & 18,994 & 2,440 & 5,224 & 26,658 \\
Flood & 3,401 & 587 & 1,315 & 5,303 & Not humanitarian & 24,427 & 3,099 & 9,145 & 36,671 \\
Hurricane & 4,517 & 651 & 1,518 & 6,686 & Rescue volunteering or donation effort & 2,270 & 344 & 680 & 3,294 \\
Landslide & 1,065 & 168 & 331 & 1,564 & \textbf{Total} & 49,353 & 6,157 & 15,688 & 71,198 \\ \cmidrule{6-10}
Not disaster & 24,459 & 3,141 & 8,885 & 36,485 & \multicolumn{5}{c}{\cellcolor[HTML]{DAE8FC}\textbf{Damage Severity}} \\ \cmidrule{6-10}
Other disaster & 1,819 & 344 & 1,154 & 3,317 & Little or none & 28,314 & 3,613 & 10,252 & 42,179 \\
\textbf{Total} & 49,353 & 6,157 & 15,688 & 71,198 & Mild & 3,904 & 698 & 1,527 & 6,129 \\ \cmidrule{1-5}
\multicolumn{5}{c|}{\cellcolor[HTML]{DAE8FC}\textbf{Informativeness}} & Severe & 17,135 & 1,846 & 3,909 & 22,890 \\ \cmidrule{1-5}
Informative & 28,073 & 3,478 & 7,206 & 38,757 & \textbf{Total} & 49,353 & 6,157 & 15,688 & 71,198 \\
Not informative & 21,280 & 2,679 & 8,482 & 32,441 &  & \multicolumn{1}{l}{} & \multicolumn{1}{l}{} & \multicolumn{1}{l}{} & \multicolumn{1}{l}{} \\
\textbf{Total} & 49,353 & 6,157 & 156,88 & 71,198 &  & \multicolumn{1}{l}{} & \multicolumn{1}{l}{} & \multicolumn{1}{l}{} & \multicolumn{1}{l}{} \\ \bottomrule
\end{tabular}
}
\caption{Annotated dataset with data splits for different tasks.}
\label{tab:task_wise_data_split}
\end{table}

\begin{table}[]
\centering
\scalebox{0.78}{
\setlength{\tabcolsep}{2.5pt}
\begin{tabular}{@{}lrrrr|rrrr@{}}
\toprule
\multicolumn{5}{c}{\textbf{Disaster Types}} & \multicolumn{4}{|c}{\textbf{Humanitarian}} \\ \midrule
\multicolumn{1}{@{}l}{\textbf{\# Labels}} & \multicolumn{1}{r}{\textbf{Train}} & \multicolumn{1}{r}{\textbf{Dev}} & \multicolumn{1}{r}{\textbf{Test}} & \multicolumn{1}{r|}{\textbf{Total}} & \multicolumn{1}{r}{\textbf{Train}} & \multicolumn{1}{r}{\textbf{Dev}} & \multicolumn{1}{r}{\textbf{Test}} & \multicolumn{1}{r@{}}{\textbf{Total}} \\ \midrule
% 1 & 32,209 & 3,609 & 9,634 & 45,452 & 1 & 40,885 & 4,777 & 11,749 & 57,411 \\
% 2 & 5,546 & 662 & 1,202 & 7,410 & 2 & 5,550 & 491 & 1,019 & 7,060 \\
% 3 & 578 & 88 & 133 & 799 & 3 & 445 & 37 & 85 & 567 \\
% 4 & 25 & 1 & 1 & 27 & \textbf{Total} & 46,880 & 5,305 & 12,853 & 65,038 \\
% 5 & 1 & 0 & 0 & 1 &  &  &  &  &  \\
% \textbf{Total} & 38,359 & 4,360 & 10,970 & 53,689 &  &  &  &  &  \\
1 & 32,227 & 3,610 & 9,635 & 45,472 & 40,885 & 4,777 & 11,749 & 57,411 \\
2 & 5,553 & 662 & 1,202 & 7,417 & 5,550 & 491 & 1,019 & 7,060 \\
3 & 579 & 88 & 133 & 800 & 445 & 37 & 85 & 567 \\ 
\textbf{Total} & \textbf{38,359} & \textbf{4,360} & \textbf{10,970} & \textbf{53,689}  & \textbf{46,880} & \textbf{5,305} & \textbf{12,853} & \textbf{65,038}
 \\
\bottomrule
\end{tabular}
}
\caption{Multilabel annotated dataset with data splits for different tasks.}
\label{tab:multilabel_task_wise_data_split}
\end{table}

\subsubsection{Crowdsourcing Results}
\label{ssec:crowdsourcing_results}
To measure the quality of the annotation, we compute the annotation agreement using Fleiss kappa~\cite{fleiss2013statistical}, Krippendorff’s alpha~\cite{krippendorff1970estimating} and average observed agreement~\cite{fleiss2013statistical}.
In Table~\ref{tab:annotation_agreement}, we present the annotation agreement for all events with different approaches mentioned above. The agreement score varies from $46\%$ to $71\%$ for different tasks. Note that, in the Kappa measurement, the values of ranges $0.41$-$0.60$, $0.61$-$0.80$, and $0.81$-$1$ refers to moderate, substantial, and perfect agreement, respectively~\cite{landis1977measurement}. Based on these measurements, we conclude that our annotation agreement score leads to moderate to substantial agreement. The number of labels and subjectivity of the annotation tasks reflected the annotation agreement score. Some annotation tasks are highly subjective. For example, for the disaster-type task, hurricane or tropical cyclones often leads to heavy rain, which causes flood (e.g., an image showing a fallen tree with flood water) can be annotated as hurricane or flood. Another example is an image showing building damage and rescue effort. In such cases, the annotation task was to carefully check what is more visible in the image and select the label accordingly. Note that, the agreement score for disaster types is comparatively lower than other tasks, which is due to the high level of subjectivity in the annotation task. Annotators needed to choose one label among seven labels.
% also reflects the complexity of the tasks in terms of the number of class labels. 
The average agreement scores are comparatively higher as we made sure at least two annotators agree on a label.

\rev{
\subsubsection{Multilabel Annotation}
\label{sssec:multilabel_annotation}
For the multilabel annotation for \textit{disaster types} and \textit{humanitarian} tasks, we followed a weak supervision approach to assign multiple labels due to the annotation budget (e.g., time, cost). We selected and assigned a \textit{set} of labels from all annotators. % mentioned above. 
Given that we have three annotators $A_{1}$, $A_{2}$, and $A_{3}$, who assigned a label $l$ from  $\mathbb{L}=\{l_{1}, l_{2}...l_{n}\}$ to an image $\mathbb{I}$, the final label set for the image $\mathbb{I}$ is defined as $\mathbb{I}_\mathbb{L}=\mathbb{S}\{A_{1}^{l},A_{1}^{l},A_{1}^{l}\}$. Here, the label with majority agreement ($\geqslant$ 66\%) is the same label as in our multiclass setting, and the rest of the labels can have a lower agreement. Note that, we were able to assign multiple labels on 53,683 images (75.4\%) for disaster types and 65,038 (91.3\%) for humanitarian tasks out of 71,198 images (see Table \ref{tab:multilabel_task_wise_data_split}). As images have been labeled in different phases and curated from existing sources, we could not properly manage to have multiple labels for all images.
}

\subsubsection{Resulting Dataset} After completing the annotation task, the proposed dataset added 155,899 labels for four tasks in addition to the existing 128,893 labels from 71,198 images. In total, this research re-annotated 65,640 images to create the MEDIC dataset. 
\rev{Furthermore, we enriched the MEDIC dataset by separately providing multilabel annotations for \textit{disaster types} and \textit{humanitarian} tasks. The distributions for multiclass and multilabel annotations are shown in Tables \ref{tab:task_wise_data_split} and \ref{tab:multilabel_task_wise_data_split}, respectively.}
\rev{We have analyzed the dataset to understand how tasks and the labels are associated with each other, for which we have computed confusion matrices between pairs of tasks. We find a good correlation between labels across tasks. For example, between humanitarian and damage severity tasks, majority of the \textit{not-humanitarian} images are also labeled as \textit{little or none} as shown in Figure \ref{fig:contingency_table_q4_q5} in Appendix \ref{ssec:appendix:data_analysis}. We have similar observations for other task pairs as well. 
}
\rev{
As for the multilabel annotation, majority of the images are labeled with single label. For example, for disaster types 84.7\% images are labeld with single label and 15.3\% with 2-3 labels. For humanitarian, 88.3\% are with single label and rest are 11.7\%. 
}

% \rev{
% We analyze most frequent label combinations for different tasks. For the humanitarian task the top most two labels are \textit{affected injured or dead people} and \textit{infrastructure and utility damage} and top most three labels are \textit{affected injured or dead people} \textit{infrastructure and utility damage} \textit{rescue volunteering or donation effort}. For the disaster type task, the top most two labels are  
% }

%% hurricane -- flood
% https://aidr-dev2.qcri.org/apps/crisismmd_images/hurricane_maria/23_9_2017/911561790168948736_0.jpg
% https://aidr-dev2.qcri.org/apps/crisismmd_images/hurricane_harvey/27_8_2017/901805267989475328_0.jpg
% https://aidr-dev2.qcri.org/apps/crisismmd_images/hurricane_maria/20_9_2017/910552318709567488_0.jpg
%https://aidr-dev2.qcri.org/apps/crisismmd_images/hurricane_maria/20_9_2017/910617882681397249_0.jpg
% https://aidr-dev2.qcri.org/apps/crisismmd_images/hurricane_maria/21_9_2017/910687961712381953_0.jpg

% \paragraph{Annotation Agreement}
\rev{
\subsection{Comparison with Other Datasets}
\label{ssec:appendix:multitask_dataset_ds}
A comparative analysis with prior disaster-related datasets suggests that the MEDIC dataset is larger in size, covering aligned labels for four tasks, and containing multilabel annotations. In Table \ref{tab:multitask_learning_datasets_dict}, we present a comparison of the datasets containing aligned labels for MTL. From the table, it is clear that the prior datasets are not designed for this kind of learning setup and the distribution of the class labels is highly skewed (see Table 9 in \cite{alam2021social} for Crisis Benchmark Dataset).
% The last row represents, MEDIC, the dataset we propose in this study, which have labels for all tasks and labels for 71,198 images. 
}
\begin{table}[!htb]
\centering
\setlength{\tabcolsep}{2.5pt}
\scalebox{0.85}{
\begin{tabular}{@{}lcccccr@{}}
\toprule
\multicolumn{1}{c}{\textbf{}} & \textbf{DT} & \textbf{Info} & \textbf{Hum} & \textbf{DS} & \textbf{Multilabel}& \multicolumn{1}{r@{}}{\textbf{Total}} \\ \midrule
CrisisMMD~\cite{alam2018crisismmd} &  & \faCheck & \faCheck & \faCheck & & 3,533 \\
Crisis Benchmark Dataset~\cite{FAlam:ASONAM20} & \faCheck & \faCheck & \faCheck & \faCheck & & 5,558 \\
\textbf{MEDIC} & \faCheck & \faCheck & \faCheck & \faCheck & \faCheck &  71,198 \\ \bottomrule
\end{tabular}
}
\caption{Multitask learning datasets for disaster image classification tasks. DT: Disaster types, Info: Informativeness, Hum: Humanitarian, DS: Damage severity. }
\label{tab:multitask_learning_datasets_dict}
\end{table}

%% file: sections/experiments.tex
\section{Experiments and Results}
\label{sec:experiments}

% \subsubsection{Statistics}
In Table \ref{tab:task_wise_data_split}, we present the dataset with task-wise data splits and distribution for multiclass setting. The distribution for multiclass setting consists of 69\%, 9\%, and 22\% for training, development, and test sets, respectively. We first conduct a baseline experiment, followed by a single-task learning experiment to compare and provide a benchmark for a multi-task setting. 

To measure the performance of each classifier and for each task setting, we use weighted average precision (P), recall (R), and F1-score (F1), which are widely used in the literature. \rev{
For the multilabel experiments we computed micro average precision (P), recall (R),  F1-score (F1) and humming loss, which are commonly used metrics~\cite{wu2017unified,sorower2010literature}. 
% In addition, we also computed weighted average P, R and F1 to compare with single-task multiclass experiments. We report such results in Appendix \ref{sec:appendix:error_analysis}.
}

\subsection{Baseline}
\label{ssec:baseline}
For the baseline experiment we evaluate \em{(i)} a majority class baseline, and \em{(ii)} fixed features from a pre-trained model used for training and testing SVM and KNN. We extracted features from the penultimate layer of the EfficientNet (b1) model~\cite{tan2019efficientnet}, which is trained using ImageNet. The majority class baseline predicts the label based on the most frequent label in the training set. This has been most commonly used in shared tasks \cite{clef-checkthat:2021:LNCS}. For training SVM and KNN we used standard parameter settings available in \textit{sci-kit learn}~\cite{pedregosa2011scikit}. 

\subsection{Single-Task Learning}
\label{ssec:single_task_learning}
We used several pre-trained models for single-task learning and fine-tuned the network with the task-specific classification layer on top of the network. This approach has been popular and has been performing well for various downstream visual recognition tasks~\cite{yosinski2014transferable,sharif2014cnn,ozbulak2016transferable,oquab2014learning}. The network architectures that we used in this study include ResNet18, ResNet50, ResNet101~\cite{he2016deep}, VGG16~\cite{simonyan2014very},  DenseNet~\cite{huang2017densely}, SqueezeNet~\cite{i2016squeezenet}, MobileNet~\cite{howard2017mobilenets}, and EfficientNet~\cite{tan2019efficientnet}. We have chosen such diverse architectures to understand their relative performance and inference time. For fine-tunning, we use the weights of the networks pre-trained using ImageNet~\cite{deng2009imagenet} to initialize our model. Our classification settings comprised binary (i.e., informativeness task) and multiclass settings (i.e., remaining three tasks). We train the models using the Adam optimizer~\cite{kingma2014adam} with an initial learning rate of $10^{-3}$, which is decreased by a factor of 10 when accuracy on the dev set stops improving for 10 epochs. The models were trained for 150 epochs. 
% \textcolor{cyan}{
We use the model with the best accuracy on the validation set to evaluate its performance on the test set.
% }

\subsection{Multi-Task Learning}
\label{ssec:multitask_learning}
% The benefits of multi-task settings can be twofold: {\em (i)} learning shared representation can help the model generalize better and improve performance on individual tasks, and {\em (ii)} training a single model instead of four different models will yield a significant speed and reduce computational load during training and inference. 
In the MEDIC dataset, the tasks share similar properties; hence, we designed a simpler approach. We use the hard parameter sharing approach to reduce the computational complexity. All tasks share the same feature layers in the network, which is followed by task-specific classification layers. For optimizing the loss, we provide equal weight to each task. Assuming that the task-specific weight is $w_i$ and task-specific loss function is $\mathcal{L}_i$, the optimization objective of the MTL is defined as $\mathcal{L}_{MTL}=\sum_{i}w_{i}.\mathcal{L}_i$. During optimization (i.e., using stochastic gradient descent to minimize the objective), the network weights in the shared layers $W_{sh}$ are updated using the following equation: 
\begin{equation}
\mathcal{W}_{sh}=\sum_{i}W_{sh}-\lambda \sum_{i}w_{i}\tfrac{\partial \mathcal{L}_{i}}{\partial W_{sh}}
\vspace{-0.5em}
\end{equation}
% \textcolor{cyan}{
%We set $w_i=1$ in our experiments. FA: i commented this as this will raise question about the other layers
We set $w_i=1$ in our experiments for all task-specific weights, i.e., equal weight for all tasks. We use softmax activation to get probability distribution over individual tasks and use cross-entropy as a loss function. We initialized the weights using pre-trained models mentioned above, which are trained using ImageNet.
% } 
Our implementation of multi-task learning supports all the network architectures mentioned in Section \ref{ssec:single_task_learning}. Therefore, we have run experiments using the same pre-trained models and same hyper-parameter settings for the MTL experiments. 
We used the NVIDIA Tesla V100-SXM2-16 GB GPU machines consisting of 12 cores and 40GB CPU memory for all experiments.

\begin{table}[]
\centering
% \resizebox{\textwidth}{!}{%
\setlength{\tabcolsep}{2.5pt}
\scalebox{0.90}{
\begin{tabular}{@{}l|cccc|cccc@{}}
\toprule
\multicolumn{1}{@{}l|}{\textbf{Model}} & \multicolumn{1}{c}{\textbf{Acc}} & \multicolumn{1}{c}{\textbf{P}} & \multicolumn{1}{c}{\textbf{R}} & \multicolumn{1}{c}{\textbf{F1}} & \multicolumn{1}{|c}{\textbf{Acc}} & \multicolumn{1}{c}{\textbf{P}} & \multicolumn{1}{c}{\textbf{R}} & \multicolumn{1}{c@{}}{\textbf{F1}} \\ \midrule
\multicolumn{1}{c|}{\textbf{}} & \multicolumn{4}{c|}{\textbf{Disaster Types}} & \multicolumn{4}{c}{\textbf{Informative}} \\\midrule
Majority & 56.6 & 32.1 & 56.6 & 41.0 & 45.9 & 21.1 & 45.9 & 28.9 \\
Eff. Net Feat. + KNN & 71.1 & 72.2 & 71.1 & 70.1 & 80.4 & 80.3 & 80.4 & 80.3 \\
Eff. Net Feat. + SVM & 75.7 & 74.1 & 75.7 & \textbf{73.2} & 83.0 & 83.0 & 83.0 & \textbf{83.0} \\ \midrule
\multicolumn{1}{c}{\textbf{}} & \multicolumn{4}{c|}{\textbf{Humanitarian}} & \multicolumn{4}{c}{\textbf{Damage Severity}} \\\midrule
Majority & 58.3 & 34.0 & 58.3 & 42.9 & 65.3 & 42.7 & 65.3 & 51.7 \\
Eff. Net Feat. + KNN & 75.3 & 74.8 & 75.3 & 74.6 & 76.5 & 73.9 & 76.5 & 74.8 \\
Eff. Net Feat. + SVM & 77.9 & 76.1 & 77.9 & \textbf{76.1} & 78.3 & 75.1 & 78.3 & \textbf{75.1} \\ \bottomrule
\end{tabular}
}
\caption{Baseline classification results. Eff. Net Feat.: Feature extracted from the penultimate layer of a pre-trained efficient net model.}
\label{tab:single_task_baseline}
\end{table}

\begin{table}[]
\centering
\scalebox{0.82}{
\setlength{\tabcolsep}{2.5pt}   

\begin{tabular}{@{}l|cccc|cccc|cccc|cccc@{}}
\toprule
\multicolumn{1}{@{}l}{\textbf{Model}} & \multicolumn{1}{|c}{\textbf{Acc}} & \multicolumn{1}{c}{\textbf{P}} & \multicolumn{1}{c}{\textbf{R}} & \multicolumn{1}{c}{\textbf{F1}} & \multicolumn{1}{|c}{\textbf{Acc}} & \multicolumn{1}{c}{\textbf{P}} & \multicolumn{1}{c}{\textbf{R}} & \multicolumn{1}{c}{\textbf{F1}} & \multicolumn{1}{|c}{\textbf{Acc}} & \multicolumn{1}{c}{\textbf{P}} & \multicolumn{1}{c}{\textbf{R}} & \multicolumn{1}{c}{\textbf{F1}} & \multicolumn{1}{|c}{\textbf{Acc}} & \multicolumn{1}{c}{\textbf{P}} & \multicolumn{1}{c}{\textbf{R}} & \multicolumn{1}{c@{}}{\textbf{F1}} \\ \midrule
\multicolumn{1}{c}{\textbf{}} & \multicolumn{8}{|c}{\cellcolor[HTML]{FFCCC9}\textbf{Disaster Types}} & \multicolumn{8}{|c}{\cellcolor[HTML]{FFCE93}\textbf{Humanitarian}} \\
\multicolumn{1}{c}{\textbf{}} & \multicolumn{4}{|c}{\textbf{Single-task}} & \multicolumn{4}{|c}{\textbf{Multi-task}} & \multicolumn{4}{c}{\textbf{Single-task}} & \multicolumn{4}{|c}{\textbf{Multi-task}} \\ \midrule
ResNet18 & 79.8 & 78.3 & 79.8 & 78.1 & 79.8 & 79.1 & 79.8 & 77.9 & 82.6 & 81.6 & 82.6 & 81.9 & 83.2 & 82.0 & 83.2 & 82.2 \\
ResNet50 & 80.6 & 79.7 & 80.6 & 79.0 & 80.9 & 80.0 & 80.9 & \textbf{79.4} & 83.4 & 83.1 & 83.4 & 83.0 & 84.2 & 83.5 & 84.2 & 83.7 \\
ResNet101 & 81.3 & 80.4 & 81.3 & 79.6 & 81.1 & 81.0 & 81.1 & 78.9 & 83.9 & 83.1 & 83.9 & 83.4 & 84.6 & 83.7 & 84.6 & 83.9 \\
VGG16 & 80.0 & 78.5 & 80.0 & 78.1 & 80.7 & 80.8 & 80.7 & 78.7 & 83.6 & 82.7 & 83.6 & 83.0 & 84.1 & 83.1 & 84.1 & 83.4 \\
DenseNet (121) & 81.1 & 80.2 & 81.1 & 79.5 & 80.7 & 80.2 & 80.7 & 78.8 & 83.4 & 82.5 & 83.4 & 82.7 & 83.9 & 83.0 & 83.9 & 83.2 \\
SqueezeNet & 76.5 & 75.0 & 76.5 & 73.6 & 77.1 & 75.5 & 77.1 & 74.7 & 79.8 & 78.0 & 79.8 & 78.4 & 81.0 & 79.5 & 81.0 & 79.9 \\
MobileNet (v2) & 80.1 & 79.0 & 80.1 & 78.0 & 79.9 & 79.2 & 79.9 & 78.0 & 82.7 & 81.7 & 82.7 & 82.0 & 83.5 & 82.5 & 83.5 & 82.7 \\
EfficientNet (b1) & 82.1 & 81.6 & 82.1 & \textbf{80.7} & 81.4 & 81.1 & 81.4 & \textbf{79.8} & 84.3 & 83.9 & 84.3 & \textbf{84.0} & 84.6 & 84.2 & 84.6 & \textbf{84.3} \\
EfficientNet (b7) & 81.0 & 79.9 & 81.0 & 79.1  & 80.5 & 79.5 & 80.5 & 78.7 & 83.2 & 82.4 & 83.2 & 82.7 & 83.2 & 82.4 & 83.2 & 82.7 \\ \midrule
\multicolumn{1}{c}{\textbf{}} & \multicolumn{8}{|c}{\cellcolor[HTML]{FFFFC7}\textbf{Informative}} & \multicolumn{8}{c}{\cellcolor[HTML]{DAE8FC}\textbf{Damage Severity}} \\ \midrule
ResNet18 & 85.9 & 86.2 & 85.9 & 85.9 & 86.8 & 86.9 & 86.8 & 86.8 & 81.4 & 78.4 & 81.4 & 79.1 & 81.7 & 78.9 & 81.7 & 79.3 \\
ResNet50 & 87.4 & 87.4 & 87.4 & \textbf{87.4} & 87.8 & 88.0 & 87.8 & 87.8 & 82.1 & 79.2 & 82.1 & 79.9 & 82.8 & 80.3 & 82.8 & 80.7 \\
ResNet101 & 87.4 & 87.6 & 87.4 & \textbf{87.4} & 88.3 & 88.3 & 88.3 & 88.3 & 82.3 & 79.9 & 82.3 & 80.6 & 82.9 & 79.9 & 82.9 & 80.2 \\
VGG16 & 86.7 & 87.1 & 86.7 & 86.8 & 87.6 & 87.7 & 87.6 & 87.6 & 82.3 & 79.6 & 82.3 & 79.7 & 82.7 & 80.1 & 82.7 & 80.5 \\
DenseNet (121) & 87.1 & 87.2 & 87.1 & \textbf{87.1} & 87.5 & 87.6 & 87.5 & 87.5 & 82.4 & 80.0 & 82.4 & 80.4 & 82.5 & 79.6 & 82.5 & 80.3 \\
SqueezeNet & 83.9 & 84.2 & 83.9 & 83.9 & 85.0 & 85.1 & 85.0 & 85.0 & 79.7 & 76.5 & 79.7 & 76.5 & 80.5 & 76.7 & 80.5 & 77.5 \\
MobileNet (v2) & 86.2 & 86.4 & 86.2 & 86.3 & 86.7 & 87.0 & 86.7 & 86.8 & 81.7 & 78.4 & 81.7 & 78.9 & 82.1 & 79.3 & 82.1 & 79.7 \\
EfficientNet (b1) & 87.7 & 87.7 & 87.7 & \textbf{87.7} & 88.6 & 88.7 & 88.6 & 88.6 & 82.8 & 80.3 & 82.8 & 80.4 & 82.9 & 80.7 & 82.9 & 80.8 \\
EfficientNet (b7) & 87.2 & 87.2 & 87.2 & \textbf{87.2} & 87.5 & 87.6 & 87.5 & 87.5 & 81.9 & 79.2 & 81.9 & 80.0 & 82.0 & 79.5 & 82.0 & 80.3 \\ \bottomrule
\end{tabular}
}
\caption{Classification results using single and multi-task settings along with different pre-trained models. Best F1 scores are highlighted.}
\label{tab:single-multitask-results}
\end{table}

\rev{
\subsection{Multilabel Classification}
\label{ssec:multilabel_classification}
In Table \ref{tab:multilabel_task_wise_data_split}, we report the distribution of multilabel data split. It shows that a major part of the dataset is labeled with a single label for both tasks. For the multilabel classification, we run experiments in a single task learning setup using the models mentioned above. We used the same training environment as other settings discussed in previous sections. However, we used sigmoid activation for multilabel instead of softmax, which is commonly used for multilabel setup. 
}

\subsection{Results}
\label{ssec:results}

\subsubsection{Baseline}
In Table \ref{tab:single_task_baseline}, we provide baseline results. From the majority baseline results it is clear that imbalance distribution does not play any role. Among SVM and KNN, the former is performing better in all tasks with 0.2 to 3.3\% improvement.

\subsubsection{Single- vs. Multi-Task Results}
In Table \ref{tab:single-multitask-results}, we report the results for both single- and multi-task settings using the mentioned models. Across different models, overall, EfficientNet (b1) performs better than other models. 
% \textcolor{cyan}{
Comparing only EfficientNet (b1) results for all tasks, the multi-task setting shows better than single-task settings; although, the difference is minor and might not be significant. However, since we share the feature layers across the four tasks, model space requirement and inference time are reduced by a factor of four. The improved inference time is crucial for real-time disaster response systems as it reduces the operational cost that running individual models would incur.

\subsubsection{Multi-Task Results using Different Random Seeds}
\label{ssec:exp_random_seed}
\textcolor{black}{
In our experiment, only the weights of the last layer were initialized randomly, hence, this can result in a minor variation in the performance. We have run experiments using different random seeds with the MTL setting. In Table \ref{tab:random_seed_results_multitask}, we report results on selected models for all tasks. We observe that variation is very minor and among different models, DenseNet (121) shows relatively lower variation across tasks.
}
\begin{table}[]
\centering
\scalebox{0.8}{
\setlength{\tabcolsep}{2.0pt}
\begin{tabular}{@{}lcccc@{}}
\toprule
\multicolumn{1}{@{}l}{\textbf{Model}} & \multicolumn{1}{c}{\textbf{DT}} & \multicolumn{1}{c}{\textbf{Info}} & \multicolumn{1}{c}{\textbf{Hum}} & \multicolumn{1}{c@{}}{\textbf{DS}} \\ \midrule
ResNet101 & 79.4 $\pm$ 0.1 & 86.2 $\pm$ 0.1 & 80.7 $\pm$ 0.1 & 80.5 $\pm$ 0.1 \\
VGG16 & 79.3 $\pm$ 0.4 & 86.2 $\pm$ 0.1 & 80.6 $\pm$ 0.1 & 80.3 $\pm$ 0.2 \\
DenseNet (121) & 79.2 $\pm$ 0.1 & \textbf{86.2 $\pm$ 0.04} & \textbf{80.7 $\pm$ 0.03} & 80.4 $\pm$ 0.1 \\
EfficientNet (b1) & \textbf{79.5 $\pm$ 0.4} & 88.5 $\pm$ 0.3 & 84.2 $\pm$ 0.1 & \textbf{80.6 $\pm$ 0.3} \\  \bottomrule
\end{tabular}
}
\caption{Experiment using different random seeds in the MTL setup.}
\label{tab:random_seed_results_multitask}
\end{table}

\subsubsection{Ablation Experiments in Multi-Task Setup}
% To understand the task correlation and how they affect performance, we also run experiments with different subsets of the tasks. In Table \ref{tab:task_pair_results}, we report the results (F1) of the models trained with different tasks combinations. It shows that we have similar performances with other combinations. %, suggesting no negative transfer between tasks.

To understand the task correlation and how they affect performance, we also run experiments with different subsets of the tasks (see Table \ref{tab:task_pair_results}). We obtain similar results with other task combinations. In Table \ref{tab:task_pair_results}, we show results obtained using combination of different subset of tasks. We observe that the results remain consistent with other combinations of tasks as well. It will be an important future research avenue to explore different weighting schemes for the tasks. Regardless, our reported results can serve as a baseline for single and multi-task disaster image classification.

\begin{table}[]
\centering
\scalebox{0.80}{
\setlength{\tabcolsep}{2.5pt}
\begin{tabular}{@{}lcccc|lcccc@{}}
\toprule
\multicolumn{1}{@{}l}{\textbf{Model (task setup)}} & \multicolumn{1}{c}{\textbf{DT}} & \multicolumn{1}{c}{\textbf{Info}} & \multicolumn{1}{c}{\textbf{Hum}} & \multicolumn{1}{c}{\textbf{DS}} & \multicolumn{1}{|l}{\textbf{Model (task setup)}} & \multicolumn{1}{c}{\textbf{DT}} & \multicolumn{1}{c}{\textbf{Info}} & \multicolumn{1}{c}{\textbf{Hum}} & \multicolumn{1}{c@{}}{\textbf{DS}} \\ \midrule
DT-Info-Hum-DS & 79.8 & 88.6 & 84.3 & 80.8 & DT-DS & 80.7 &  &  & \textbf{81.3} \\
DT-Info-Hum & 80.3 & \textbf{88.9} & \textbf{84.5} &  & Info-Hum-DS &  & 88.3 & 84.0 & 80.8 \\
DT-Info-DS & 80.2 & 88.6 &  & \textbf{81.0} & Info-Hum &  & \textbf{88.5} & 83.9 &  \\
DT-Info & 80.1 & 88.7 &  &  & Info-DS &  & 88.2 &  & 80.5 \\
DT-Hum & \textbf{80.5} &  & 84.4 &  & Hum-DS &  &  & 84.1 & 80.8 \\ 
\bottomrule
\end{tabular}
}
\caption{Results (F1) with different combination of tasks using Efficient-
Net (b1). DT: Disaster type, Info: Informativeness, Hum: Humanitarian, DS:
Damage severity.}
\label{tab:task_pair_results}
\end{table}

\subsubsection{Multilabel Classification Results}
\rev{
In Table \ref{tab:multilabel_results}, we report multilabel classification results for disaster types and humanitarian tasks. Overall, across different models, SqueezeNet is the worst performing model, which we also observed for single and multi-task multiclass classification results. The multilabel results, as in Table \ref{tab:task_pair_results}, are not equally comparable with multiclass results, as reported in Table \ref{tab:single-multitask-results}. This results will serve as baselines in future studies. 
}

\begin{table}[tbh!]
\centering
\setlength{\tabcolsep}{3.0pt}
\scalebox{0.78}{
\begin{tabular}{@{}l|ccccc|ccccc@{}}
\toprule
\multicolumn{1}{@{}l|}{\textbf{Model}} & \multicolumn{1}{c}{\textbf{Acc}} & \multicolumn{1}{c}{\textbf{P}} & \multicolumn{1}{c}{\textbf{R}} & \multicolumn{1}{c}{\textbf{F1}} & \multicolumn{1}{c}{\textbf{H}} & \multicolumn{1}{|c}{\textbf{Acc}} & \multicolumn{1}{c}{\textbf{P}} & \multicolumn{1}{c}{\textbf{R}} & \multicolumn{1}{c}{\textbf{F1}} & \multicolumn{1}{c@{}}{\textbf{H}} \\ \midrule
\multicolumn{1}{c}{} & \multicolumn{5}{c}{Disaster Type} & \multicolumn{5}{c}{Humanitarian} \\ \midrule
ResNet18 & 73.9 & 86.2 & 73.2 & 79.2 & 6.2 & 76.5 & 85.4 & 78.4 & 81.8 & 9.6 \\
ResNet50 & 76.3 & 86.1 & 75.6 & 80.5 & 5.9 & 78.6 & 86.4 & 80.5 & 83.3 & 8.8 \\
ResNet101 & 75.8 & 86.2 & 75.9 & 80.7 & 5.9 & 79.0 & 86.6 & 80.4 & \textbf{83.4} & 8.8 \\
VGG16 & 76.2 & 86.8 & 75.6 & \textbf{80.8} & 5.8 & 78.9 & 86.3 & 80.5 & 83.3 & 8.8 \\
DenseNet (121) & 75.8 & 87.6 & 74.3 & 80.4 & 5.9 & 78.0 & 86.5 & 78.9 & 82.5 & 9.1 \\
SqueezeNet & 36.3 & 41.8 & 64.3 & 50.7 & 20.3 & 31.9 & 55.3 & 78.9 & 65.0 & 23.2 \\
MobileNet (v2) & 73.5 & 86.4 & 71.8 & 78.4 & 6.4 & 76.8 & 86.0 & 78.0 & 81.8 & 9.5 \\
EfficientNet (b1) & 73.4 & 86.1 & 71.5 & 78.1 & 6.5 & 77.9 & 86.1 & 79.9 & 82.9 & 9.0 \\
EfficientNet (b7) & 76.0 & 86.0 & 74.7 & 80.0 & 6.1 & 78.2 & 85.4 & 80.3 & 82.8 & 9.1 \\ \bottomrule
\end{tabular}
}
\caption{Classification results using single-task multilabel settings with different pre-trained models. \textit{H: Humming Loss} lower is better. \textit{Micro} average precision, recall, and F1.}
\label{tab:multilabel_results}
\end{table}

\begin{table}[h]
\centering
\setlength{\tabcolsep}{3.0pt}
\scalebox{0.78}{
\begin{tabular}{@{}l|ccc|ccc@{}}
\toprule
\multicolumn{1}{@{}l}{\textbf{Label}} & \multicolumn{1}{c}{\textbf{P}} & \multicolumn{1}{c}{\textbf{R}} & \multicolumn{1}{c|}{\textbf{F1}} & \multicolumn{1}{c}{\textbf{P}} & \multicolumn{1}{c}{\textbf{R}} & \multicolumn{1}{c@{}}{\textbf{F1}} \\ \midrule
\multicolumn{1}{c|}{\textbf{}} & \multicolumn{3}{c|}{\textbf{Single-task}} & \multicolumn{3}{c}{\textbf{Multi-task}} \\ \midrule
\multicolumn{7}{c}{\textbf{Disaster Type}} \\\midrule
Earthquake & 73.8 & 82.6 & \textbf{77.9} & 70.5 & 83.5 & 76.4 \\
Fire & 78.2 & 85.2 & \textbf{81.6} & 74.1 & 85.4 & 79.3 \\
Flood & 78.1 & 80.7 & 79.4 & 78.5 & 80.8 & \textbf{79.6} \\
Hurricane & 65.6 & 67.5 & \textbf{66.6} & 64.4 & 63.0 & 63.7 \\
Landslide & 62.2 & 78.5 & \textbf{69.4} & 60.4 & 75.5 & 67.1 \\
Not disaster & 88.9 & 92.9 & \textbf{90.9} & 88.9 & 92.7 & 90.8 \\
Other disaster & 70.5 & 18.8 & \textbf{29.7} & 72.6 & 15.6 & 25.7 \\ \midrule
\multicolumn{7}{c}{\textbf{Informativeness}} \\ \midrule
Informative & 86.5 & 86.8 & 86.7 & 85.8 & 90.0 & \textbf{87.9} \\
Not-informative & 88.8 & 88.5 & 88.6 & 91.2 & 87.3 & \textbf{89.2} \\ \midrule
\multicolumn{7}{c}{\textbf{Humanitarian}} \\ \midrule
Affected, injured, or dead people & 54.8 & 42.6 & \textbf{47.9} & 51.5 & 43.8 & 47.3 \\
Infrastructure and utility damage & 81.5 & 85.1 & 83.2 & 80.6 & 87.8 & \textbf{84.1} \\
Not humanitarian & 89.8 & 89.9 & 89.9 & 91.1 & 89.2 & \textbf{90.1} \\
Rescue volunteering or donation effort & 48.7 & 42.2 & \textbf{45.2} & 49.4 & 36.2 & 41.8 \\ \midrule
\multicolumn{7}{c}{\textbf{Damage Severity}} \\ \midrule
Litle or none & 89.7 & 93.2 & 91.4 & 91.0 & 92.4 & \textbf{91.7} \\
Mild & 42.3 & 9.8 & 15.9 & 40.7 & 11.7 & \textbf{18.2} \\
Severe & 70.2 & 84.3 & \textbf{76.6} & 69.2 & 85.6 & 76.5 \\ \bottomrule
\end{tabular}
}
\caption{Class-wise results for both single and multi-task settings using EfficientNet (b1) model}
\label{tab:class_wise_results}
\end{table}

\begin{table}[h]
\centering
\setlength{\tabcolsep}{2.5pt}
\scalebox{0.8}{
\begin{tabular}{@{}l|rrrr|r|r@{}}
\toprule
\multicolumn{1}{@{}l|}{\textbf{Model}} & \multicolumn{5}{c|}{\textbf{Single-task}} & \multicolumn{1}{c}{\textbf{Multi-task}} \\ \midrule
\multicolumn{1}{c}{\textbf{}} & \multicolumn{1}{|c}{\textbf{DT}} & \multicolumn{1}{c}{\textbf{Info}} & \multicolumn{1}{c}{\textbf{Hum}} & \multicolumn{1}{c}{\textbf{DS}} & \multicolumn{1}{|c}{\textbf{Sum}} & \multicolumn{1}{|c}{\textbf{All tasks}} \\ \midrule %\cline{2-7}
\multicolumn{1}{c}{\textbf{}} & \multicolumn{6}{c}{\textbf{Training time on the train set with 49,353 images}} \\ \midrule
ResNet18 & 21:38:48 & 17:15:09 & 16:53:02 & 17:41:23 & 3 days, 1:28:22 & 1 day, 3:41:02 \\
ResNet50 & 21:14:49 & 17:16:03 & 21:41:07 & 17:24:00 & 3 days, 5:35:59 & 18:19:56 \\
ResNet101 & 27:35:29 & 18:37:23 & 17:41:23 & 19:49:31 & 3 days, 11:43:46 & 1 day, 0:27:28 \\
VGG16 & 19:53:52 & 23:43:49 & 23:15:04 & 23:37:10 & 3 days, 18:29:55 & 22:41:41 \\
DenseNet (121) & 20:23:39 & 17:08:27 & 17:23:06 & 18:21:06 & 3 days, 1:16:18 & 18:20:41 \\
SqueezeNet & 24:12:26 & 17:18:55 & 20:26:42 & 16:47:46 & 3 days, 6:45:49 & 18:12:44 \\
MobileNet (v2) & 17:44:03 & 21:39:41 & 17:55:16 & 21:06:44 & 3 days, 6:25:44 & 15:53:10 \\
EfficientNet (b1) & 21:59:19 & 17:37:01 & 17:28:30 & 17:08:27 & 3 days, 2:13:17 & 20:38:06 \\
EfficientNet (b7) &  & 26:39:17 & 26:40:33 & 26:55:17 & 3 days, 8:15:07 & 1 day, 16:13:38 \\\midrule
\multicolumn{1}{c}{\textbf{}} & \multicolumn{6}{c}{\textbf{Inference time on the test set with 15,688 images}} \\ \midrule
ResNet18 & 0:02:26 & 0:01:56 & 0:05:11 & 0:01:53 & 0:11:26 & 0:05:10 \\
ResNet50 & 0:02:25 & 0:01:55 & 0:02:24 & 0:01:54 & 0:08:38 & 0:02:13 \\
ResNet101 & 0:05:20 & 0:07:48 & 0:02:05 & 0:02:08 & 0:17:21 & 0:01:58 \\
VGG16 & 0:05:21 & 0:01:57 & 0:05:10 & 0:01:56 & 0:14:24 & 0:02:15 \\
DenseNet (121) & 0:02:08 & 0:01:55 & 0:01:57 & 0:05:22 & 0:11:22 & 0:02:08 \\
SqueezeNet & 0:10:59 & 0:01:54 & 0:02:22 & 0:05:15 & 0:20:30 & 0:04:44 \\
MobileNet (v2) & 0:01:57 & 0:02:26 & 0:01:56 & 0:02:26 & 0:08:45 & 0:01:57 \\
EfficientNet (b1) & 0:05:17 & 0:01:56 & 0:02:07 & 0:01:54 & 0:11:14 & 0:02:32 \\
EfficientNet (b7) &  & 0:02:12 & 0:02:11 & 0:02:10 & 0:06:33 & 0:02:13 \\ \bottomrule
\end{tabular}
}
\caption{Training and inference time in single- vs. multi-task settings with a batch size of 32. Time is in \textit{day, hour:minute:second format}.}
\label{tab:training_inference_time_single_vs_multitask}
\end{table}

\subsubsection{Error Analysis}
Given that class distribution can play a significant role in classifier performance, we explored whether low prevalent classes have any significant impact. In Table \ref{tab:class_wise_results}, we report task-wise classification results for both single and multi-task settings in which the model is trained using EfficientNet model. It appears that low prevalent classes have lower performance. However, this is not always the case. For example, the distribution of \textit{Fire} class label is 3.8\% in the dataset but the performance is third-best among class labels. Where the distribution of \textit{Other disaster} is 5.1\%, however, the F1 is 27.0, which is the lowest performance. With our analysis, we found that this \textit{Other disaster} confused with \textit{Not disaster}.   

% \rev{TODO: comment on confusion matrix in appendix}
% In Table \ref{tab:dt_vs_info_hum_ds}, we report classification results of EfficientNet (b1) multitask learning model, which shows that disaster type class label predictions and their prediction with other tasks. The results suggests that the higher distribution of \textit{Not disaster} does not effect the classification performance much.  
\rev{
In Tables \ref{tab:conf_mat_disaster_types}, \ref{tab:conf_mat_informative}, \ref{tab:conf_mat_humanitarian} and \ref{tab:conf_mat_damage_severity} (in Appendix~\ref{sec:appendix:error_analysis}) we report classification confusion matrices using EfficientNet (b1) model for disaster types, informative, humanitarian and damage severity, respectively. From the tables, we observe that  there is comparable performances between different task settings. In some cases class label performance increases in multi-task  setting and in some cases it decreases. For example, true positives increase for informative and decreases for not-informative in multi-task setting. The results in these tables also confirm the results in Table \ref{tab:single-multitask-results}.
}

\subsubsection{Computational Time Analysis}
We have done extensive analysis to understand whether multi-task learning setup reduces computational time. In Table \ref{tab:training_inference_time_single_vs_multitask}, we provide such findings for all the models we used in our experiments. From the results, it is clear that multi-task learning setup can significantly reduce the computation time both in terms of training and inference.

%% file: sections/discussion.tex
\begin{figure}[]
\centering
\includegraphics[width=0.8\textwidth]{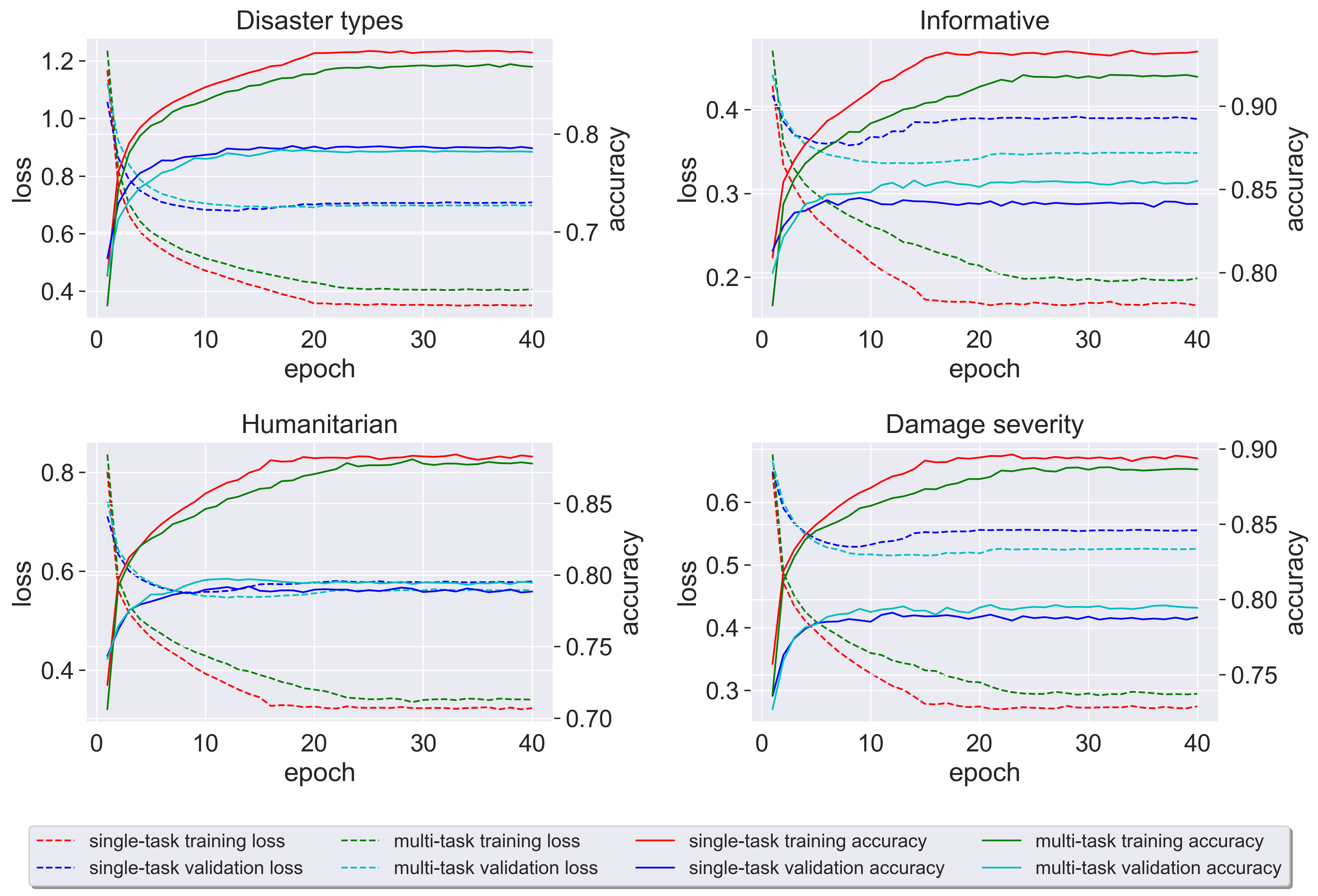}
\caption{Training and validation loss and accuracy for EfficientNet (b1) model for single and multi-task settings. }
\label{fig:train}
% \vspace{-0.5em}
\end{figure}

\section{Discussion and Future Work}
\label{sec:discussions}

The MEDIC dataset provides images from diverse events consisting of different time frames. The crowdsourced annotation provides a reasonable annotator agreement even though the task is subjective. Our experiments show that multi-task learning with neural networks reduces computational complexity significantly while having comparative performance. 

% \textcolor{cyan}{
In Figure \ref{fig:train}, we show the loss and accuracy plots for single and multi-task settings for EfficientNet (b1) model. We limit the plots to 40 epochs as all of the models converged by then. We notice similar convergence rates for both single and multi-task learning setups. We observe that the multi-task objective function acts as a regularizer as the training loss is consistently higher and training accuracy is lower than the single-task setting while having similar or better performance on the validation set. This suggests that the multi-task setup may benefit from models having a larger capacity.
% }

Class distribution is an important issue that affect classifier performance. We investigated class-wise performances and confusion matrix. Our observation suggests that imbalanced class distribution is not only factor for lower classification performance in certain classes. It also depends on distinguishing properties of the class label. For example, the distribution of \textit{Fire} class label is 3.8\% in the dataset but the performance is third-best among class labels. Where the distribution of \textit{Other disaster} is 5.1\%, however, the F1 is 27.0, which is the lowest performance. 
% In appendix Section \ref{sec:appendix:exp_settings}, Table \ref{tab:class_wise_results}, we reported class-wise results. 

% \paragraph{Limitation}
% \label{para:limitation}
% We foresee several limitations of our work. As mentioned earlier disaster types and humanitarian tasks can be annotated with multiple labels, which we annotated with the single label in this study. Even though our choice has been influenced based on the knowledge of humanitarian organizations, however, we aim to explore it further. In our experiments, we may have to explore a much larger network, which can help multi-task learning better. 
%We realized that the same image could be annotated with multiple labels such as infrastructure damage and rescue effort for humanitarian tasks. For this study, we only focused on single-label annotation to reduce the complexity of the annotation efforts.

\paragraph{Future Work}
\label{para:future_work}
Our future work includes %annotating images with multilabel annotation, 
exploring other multi-task learning methods, and investigating tasks groups and relationships. For instance, further investigation is needed to explain why training the model with disaster types, informativeness and humanitarian tasks reduces performance as presented in Table \ref{tab:task_pair_results}. Other research avenues include multimodality (e.g., integrating text), and investigating class imbalance issues. % (e.g., up-sampling, down-sampling), or utilizing unlabeled data.    

% \textcolor{blue}{Multimodality}

% \textcolor{blue}{skewed distribution issues}

%% file: sections/conclusion.tex
\section{Conclusions}
\label{sec:conclutions}
We presented a large-scale, manually annotated multi-task learning dataset, comprising 71,198 images labeled for four tasks, which were specifically designed for multi-task learning research and disaster response image classification. The dataset will not only be useful to develop robust models for disaster response tasks but will also enable evaluation of general multi-task models. We provide classification results using nine different pre-trained models, which can serve as a benchmark for future work. We report that the multi-task model reduces the inference time significantly, hence, such a model can be very useful for real-time classification tasks, especially for analyzing social media image streams.

%% file: sections/supplemental_material.tex
\section{Data Collection}
\label{appendix:data_collection}

\subsection{Data Curation and Annotation}
We extended the Crisis Benchmark dataset to develop MEDIC, a multitask learning dataset for disaster response. For the annotation, we provided detailed instructions to the annotators, which they followed during the annotation tasks. Our annotation consists of four tasks in different batches, and we provided task-specific instructions along with them.  

\subsection{Annotation Instructions}
\label{ssec:appendix:annotation_instructions}
The annotation task involves identifying images that are useful for humanitarian aid/response. During different disaster events (i.e., natural and human-induced or hybrid), \emph{humanitarian aid}\footnote{\url{https://en.wikipedia.org/wiki/Humanitarian_aid}} involves assisting people who need help. The primary purpose of humanitarian aid is to save lives, reduce suffering, and rebuild affected communities. Among the people in need belong homeless, refugees, and victims of natural disasters, wars, and conflicts who need necessities like food, water, shelter, medical assistance, and damage-free critical infrastructure and utilities such as roads, bridges, power lines, and communication poles. 

For disaster types and humanitarian tasks, it is possible that some images can be annotated with multiple labels. In such cases, the instruction is to choose a label that is critical (i.e., higher priority) for humanitarian organizations and more prominent in the image. 

\begin{figure}[h]
\centering
\includegraphics[width=1.0\textwidth]{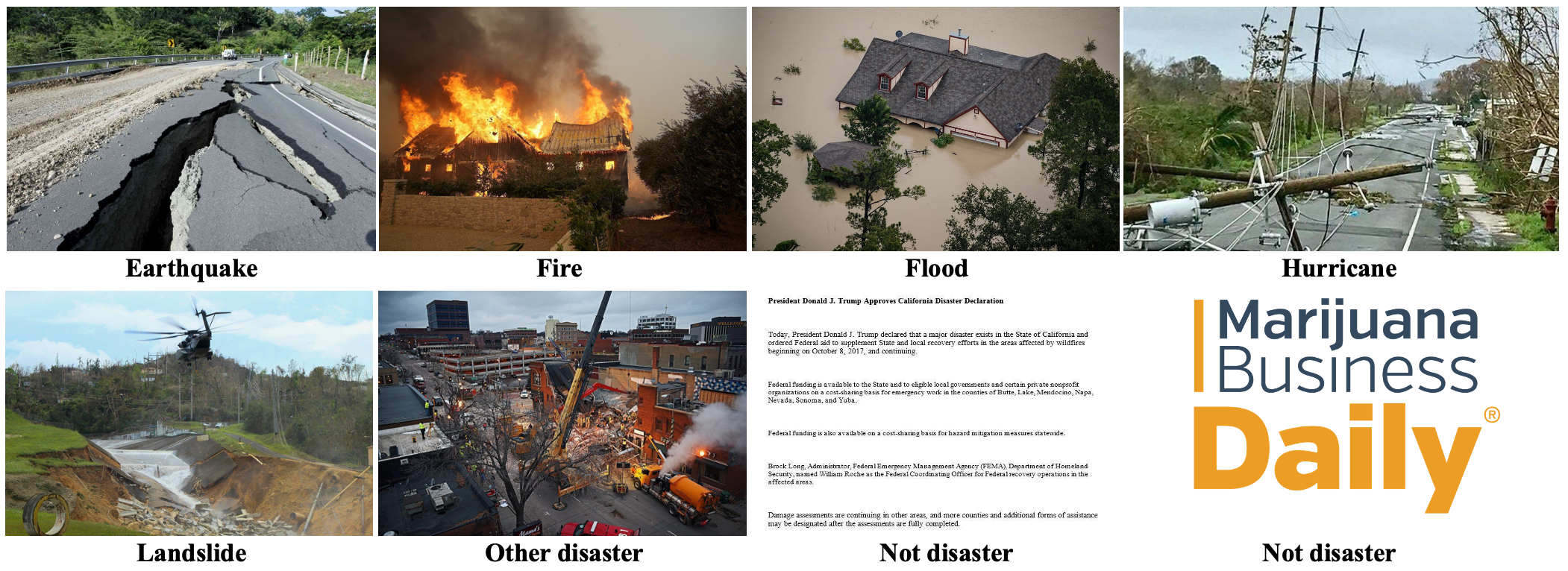}
\caption{Examples of images \textbf{disaster types}.}
\label{fig:example_disaster_types}
\end{figure}

\subsubsection{Disaster Types}
The purpose of identifying disaster type is to understand the type of disaster events shared in an image. The annotation task involves looking into the image can carefully select one of the following disaster types based on their specific definition. There might be the case that an image shows an effect of a hurricane (destroyed house) and also flood, in such cases the task is to carefully check what is more visible and select label accordingly. Example of images demonstrating different disaster types is shown in Figure \ref{fig:example_disaster_types}.

\begin{itemize}
    \item \textbf{Earthquake:} this type of images shows damaged or destroyed buildings, fractured houses, ground ruptures such as railway lines, roads, airport runways, highways, bridges, and tunnels.
    \item \textbf{Fire:} image shows man-made fires or wildfires (forests, grasslands, brush, and deserts), destroyed forests, houses, or infrastructures. 
    \item \textbf{Flood:} image shows flooded areas, houses, roads, and other infrastructures. 
    \item \textbf{Hurricane:} image shows high winds, a storm surge, heavy rains, collapsed electricity polls, grids, and trees.
    \item \textbf{Landslide:} image shows landslide, mudslide, landslip, rockfall, rockslide, earth slip, and land collapse 
    % then we label them as \textit{landslide} otherwise \textit{not landslide}. 
    \item \textbf{Other disasters:} image shows any other disaster types such as plane crash, bus, car, or train accident, explosion, war, and conflicts.
    \item \textbf{Not disaster:} image shows cartoon, advertisement, or anything that cannot be easily linked to any disaster type.
\end{itemize}

\begin{figure}[h]
  \centering
  \includegraphics[width=0.7\textwidth]{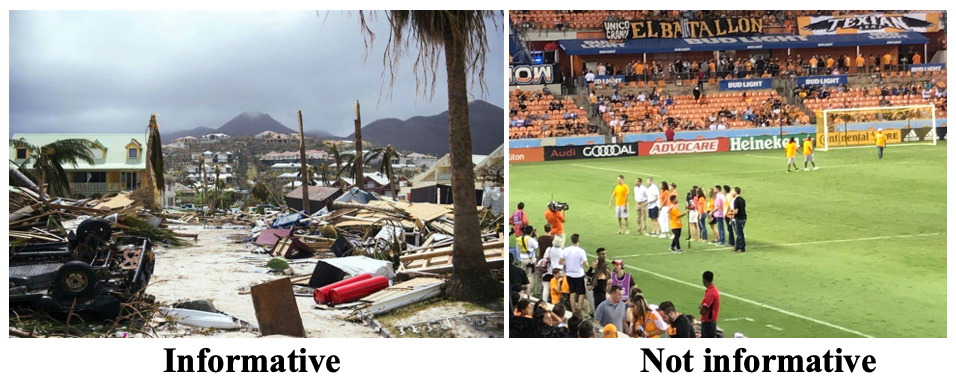}
  \caption{Example images for \textbf{informativeness}.}
  \label{fig:example_informativeness}
\end{figure}

\subsubsection{Informativeness}
The purpose of this task is to determine whether image is useful for \emph{humanitarian aid} purposes as defined below. If the given image is useful for \emph{humanitarian aid}, the annotation task is to select the label ``Informative'', otherwise select the label ``Not informative'' image.
Example of images demonstrating informative vs. not-informative is shown in Figure \ref{fig:example_disaster_types}.
\begin{itemize}
\item \textbf{Informative:} if an image is useful for humanitarian aid and shows one or more of the following: cautions, advice, and warnings, injured, dead, or affected people, rescue, volunteering, or donation request or effort, damaged houses, damaged roads, damaged buildings; flooded houses, flooded streets; blocked roads, blocked bridges, blocked pathways; any built structure affected by earthquake, fire, heavy rain, strong winds, gust, etc., disaster area maps.
\item \textbf{Not informative:} if the image is not useful for humanitarian aid and shows advertising, banners, logos, cartoons, and blurred.
\end{itemize}

\begin{figure}[h]
  \centering
  \includegraphics[width=1.0\textwidth]{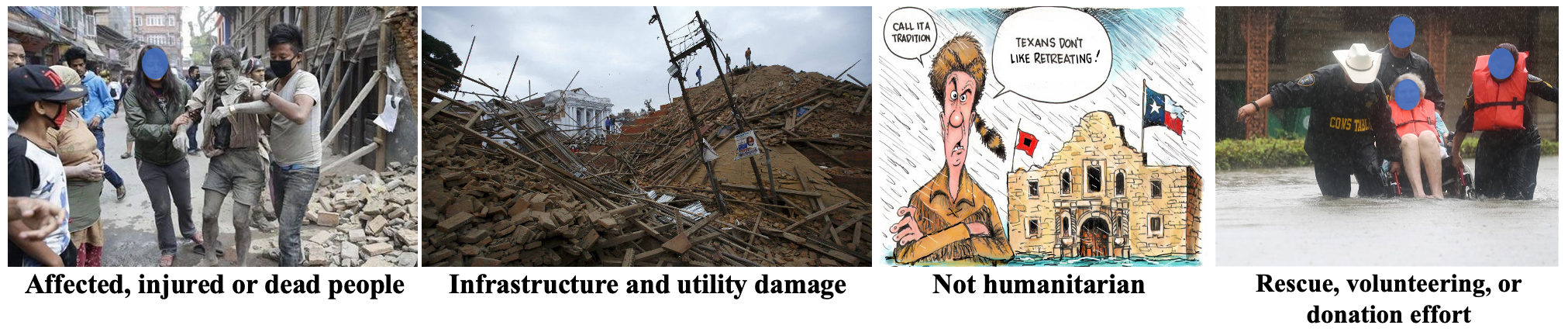}
  \caption{Example images for \textbf{humanitarian} categories.}
  \label{fig:example_humanitarian}
\end{figure}
\subsubsection{Humanitarian Categories}
Based on the \emph{humanitarian aid} definition above, we define each \textbf{humanitarian} information category below.

\begin{itemize} %[label={\bfseries L\arabic*:},leftmargin=*]
	\item \textbf{Affected, injured or dead people:} image shows injured, dead, or affected people such as people in shelter facilities, sitting or lying outside, etc.
	\item \textbf{Infrastructure and utility damage:} image shows any built structure affected or damaged by the disaster. This includes damaged houses, roads, buildings; flooded houses, streets, highways; blocked roads, bridges, pathways; collapsed bridges, power lines, communication poles, etc.
	\item \textbf{Not humanitarian:} image is not relevant or useful for humanitarian aid and response such as non-disaster scenes, cartoons, advertisement banners, celebrities, etc.
	\item \textbf{Rescue, volunteering, or donation effort:} image shows any type of rescue, volunteering, or response effort such as people being transported to safe places, people being evacuated from the hazardous area, people receiving medical aid or food, donation of money, blood, or services, etc.
\end{itemize}

\subsubsection{Damage Severity}
The purpose of this task is to identify the severity of damage reported in an image. It can be physical destruction to a build-structure. Our goal is to detect physical damages like broken bridges, collapsed or shattered buildings, destroyed or creaked roads. 
We define each damage severity category below.

\begin{enumerate}
    \item \textbf{Severe:} Substantial destruction of an infrastructure belongs to the severe damage category. For example, a non-livable or non-usable building, a non-crossable bridge, or a non-drivable road, destroyed, burned crops, forests are all examples of severely damaged infrastructures. 
    \indent For example, if one or more building in the image show substantial loss of amenity or images shows a building that is not safe to use then such image should be labeled as severe damage. 
    \item \textbf{Mild:} Partially destroyed buildings, bridges, houses, roads belong to mild damage category.
    For example, if image shows a building with damage upto 50\%, partial loss of amenity/roof or part of the building can has to be closed down then it should label as mild damage.
    \item \textbf{Little or none:} Images that show damage-free infrastructure (except for wear and tear due to age or disrepair) belong to the little-or-no-damage category.
\end{enumerate}

\begin{figure}[h]
  \centering
  \includegraphics[width=1.0\textwidth]{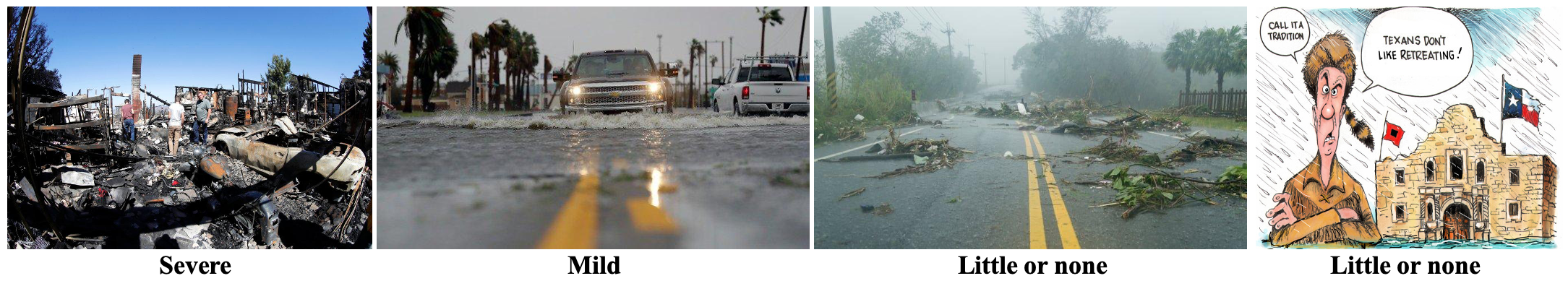}
  \caption{Example images for \textbf{damage severity}.}
  \label{fig:example_damage_severity}
\end{figure}

\begin{figure}[h]
\centering
\includegraphics[width=1.0\textwidth]{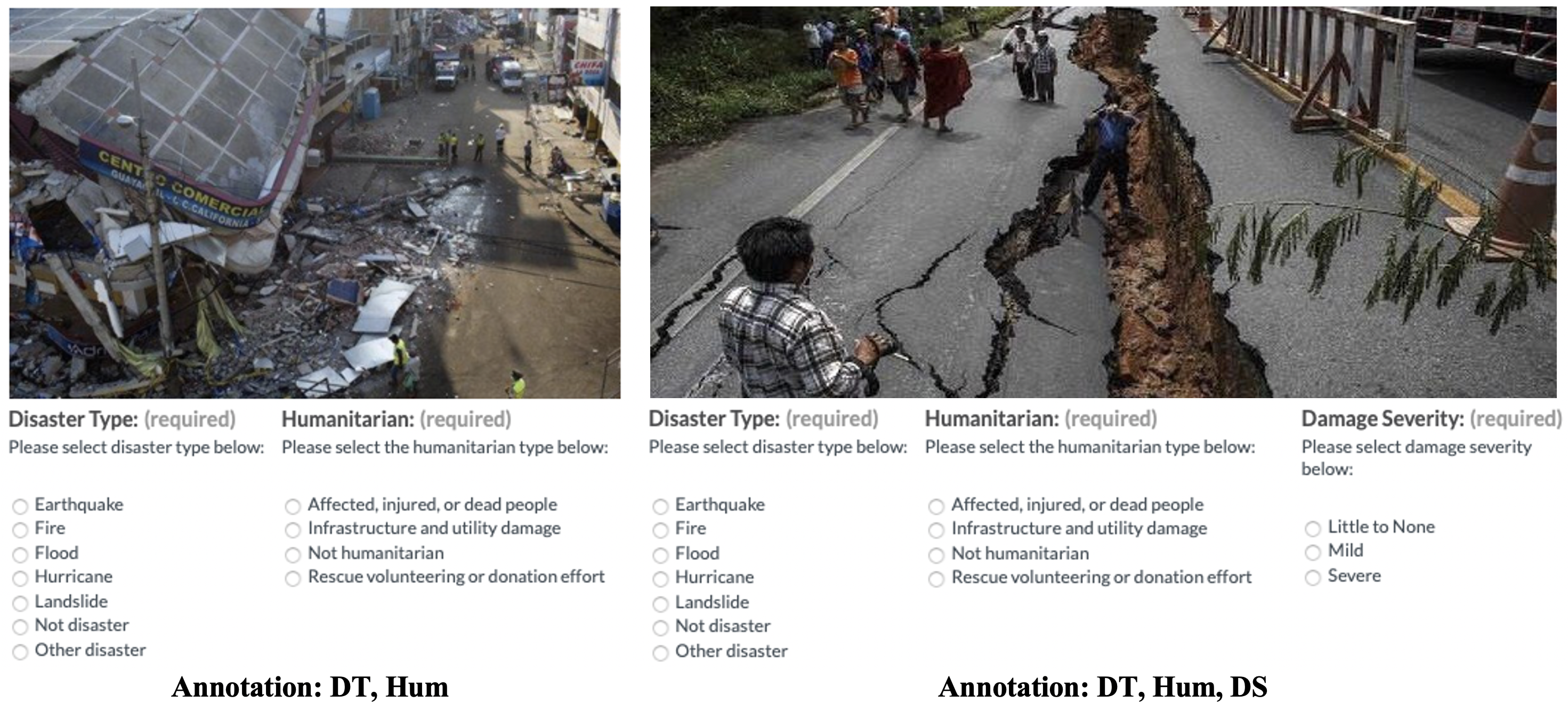}
\caption{Example of annotation interfaces on Appen crowdsoursing platform. DT: disaster type, Hum: humanitarian, DS: damage severity.}
\label{fig:example_annotation_interface}
\end{figure}

\subsection{Annotation Interface}
\label{ssec:appendix:annotation_interface}
An example of annotation interface is showin in Figure \ref{fig:example_annotation_interface}. Image on the left shows annotation task is launched to annotate image for disaster type and humanitarian tasks and image on the right shows annotation task is launched for three tasks.

\subsection{Manual Annotation}
\label{ssec:appendix:manual_annotation}
In our annotation tasks through the Appen platform, more than 3000 annotators participated from more than 50 countries. For the annotation task, we estimated hourly wages and it was 6 to 8 USD per hour on average, which varied depending on the two to three labels annotation per image. We think such pay is reasonable as annotators are from various part of the world where wages varies depending on the location. In total we paid 5,159 USD for the annotation, including Appen charges.

% %  we considered the following evaluation measures. 
% % \begin{enumerate}[leftmargin=*]
% % 	\itemsep-0.1em 
% % 	\item Fleiss kappa: It is a reliability measure that is applicable for any fixed number of annotators annotating categorical labels to a fixed number of items, which can handle two or more categories and annotators~\cite{fleiss2013statistical}. However, it can not handle missing labels, except for excluding them from the computation. 
% % 	\item Average observed agreement: It is an average observed agreement over all pairs of annotators \cite{fleiss2013statistical}.
% % 	\item Majority agreement: We compute the majority at the tweet level and take the average. The reason behind this is that for many tweets the number of annotators vary between three and five, and hence, it is plausible to evaluate the agreement at the tweet level.
% % 	\item Krippendorff’s alpha: It is a measure of agreement that allows two or more annotators and categories~\cite{krippendorff1970estimating}. Additionally, it handles missing labels. 
% % \end{enumerate}
% s

\subsection{Data Analysis}
\label{ssec:appendix:data_analysis}
\begin{figure*}
\centering
    \begin{subfigure}[b]{0.45\textwidth}
        \includegraphics[width=\textwidth]{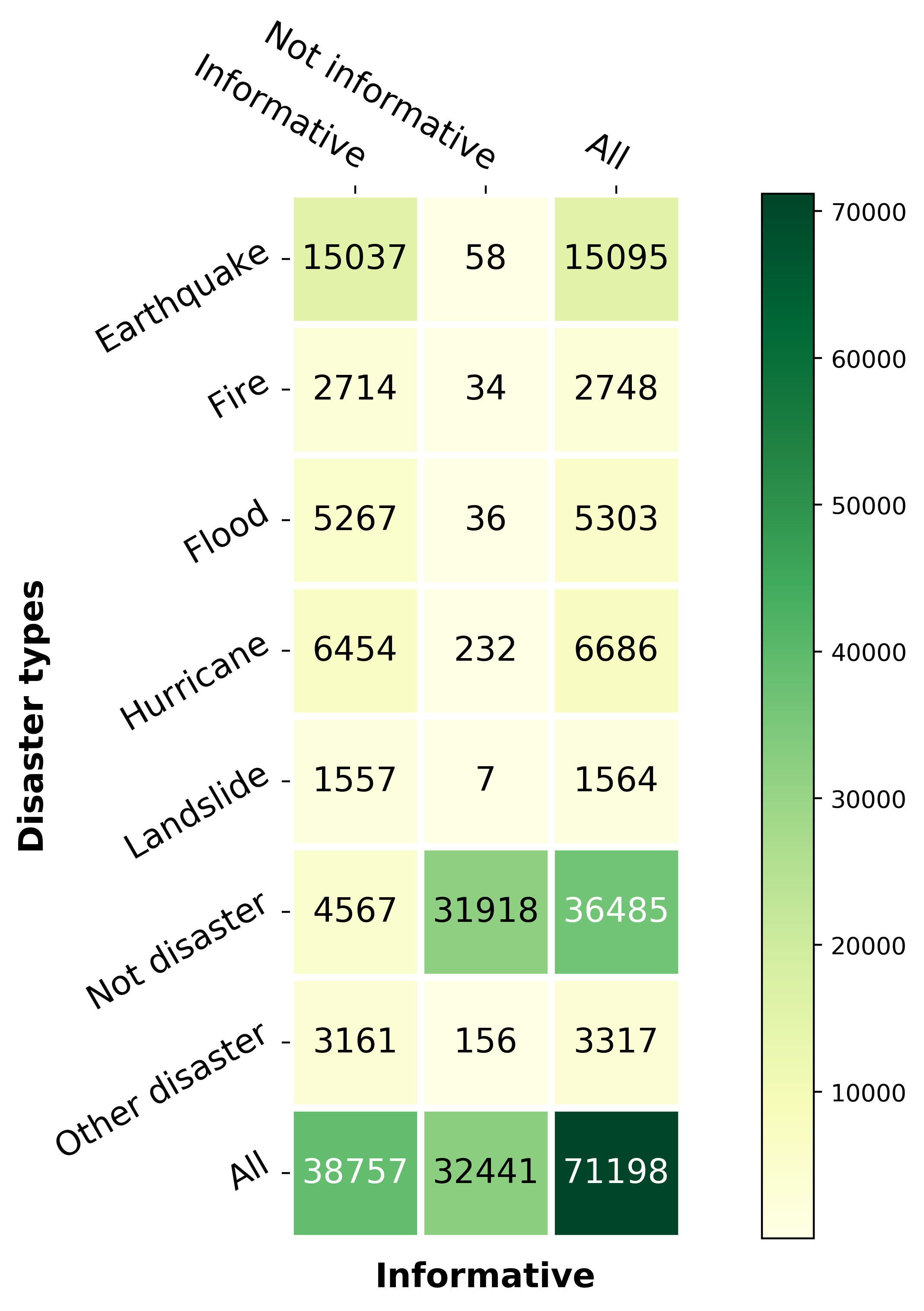}
        \caption{DT and info.}
        \label{fig:contingency_table_dt_info}
    \end{subfigure}%
    \begin{subfigure}[b]{0.45\textwidth}    
        \includegraphics[width=\textwidth]{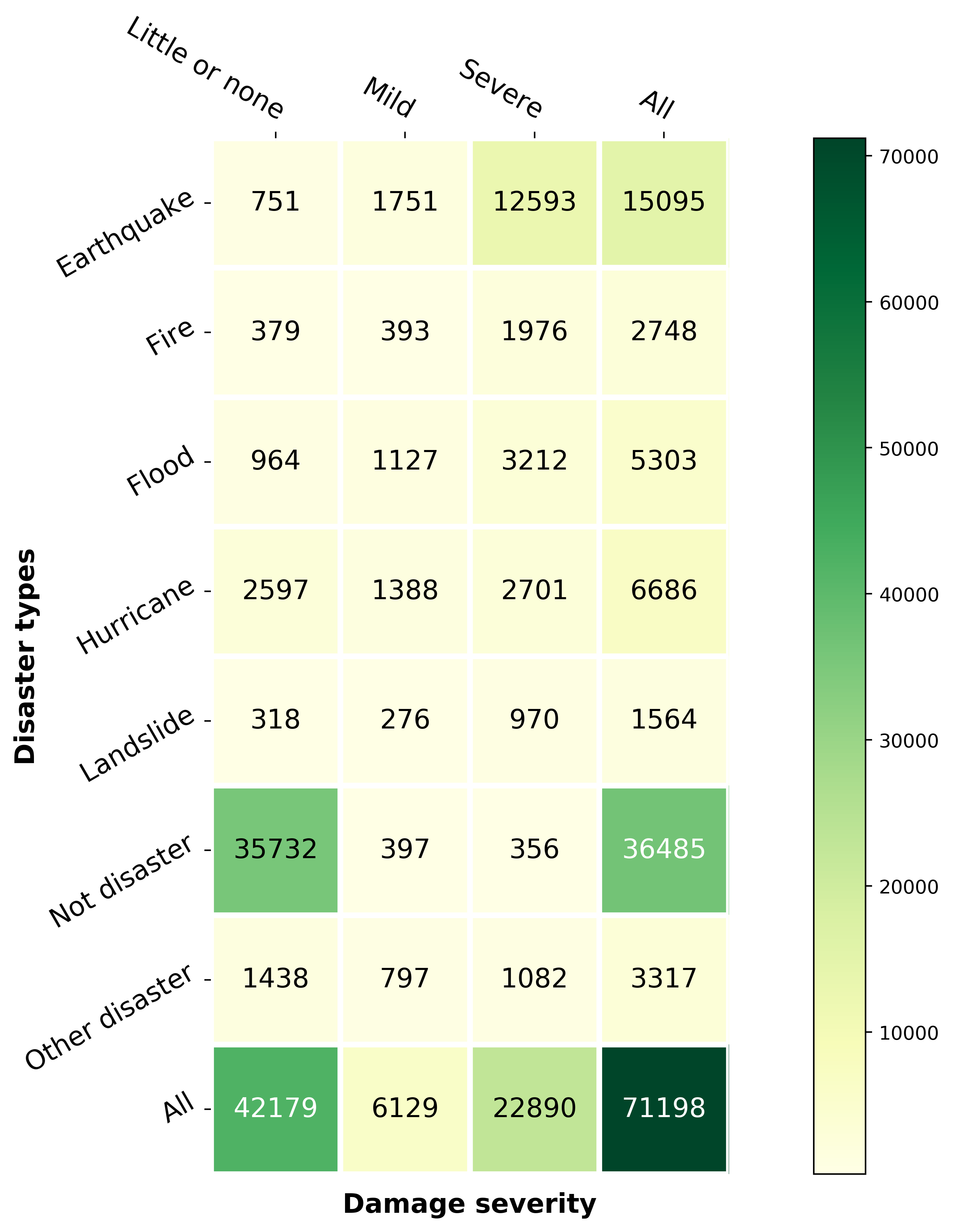}
        \caption{DT and DS.}
        \label{fig:contingency_table_dt_ds}    
    \end{subfigure}
    \hfill
    \begin{subfigure}[b]{0.45\textwidth}    
        \includegraphics[width=\textwidth]{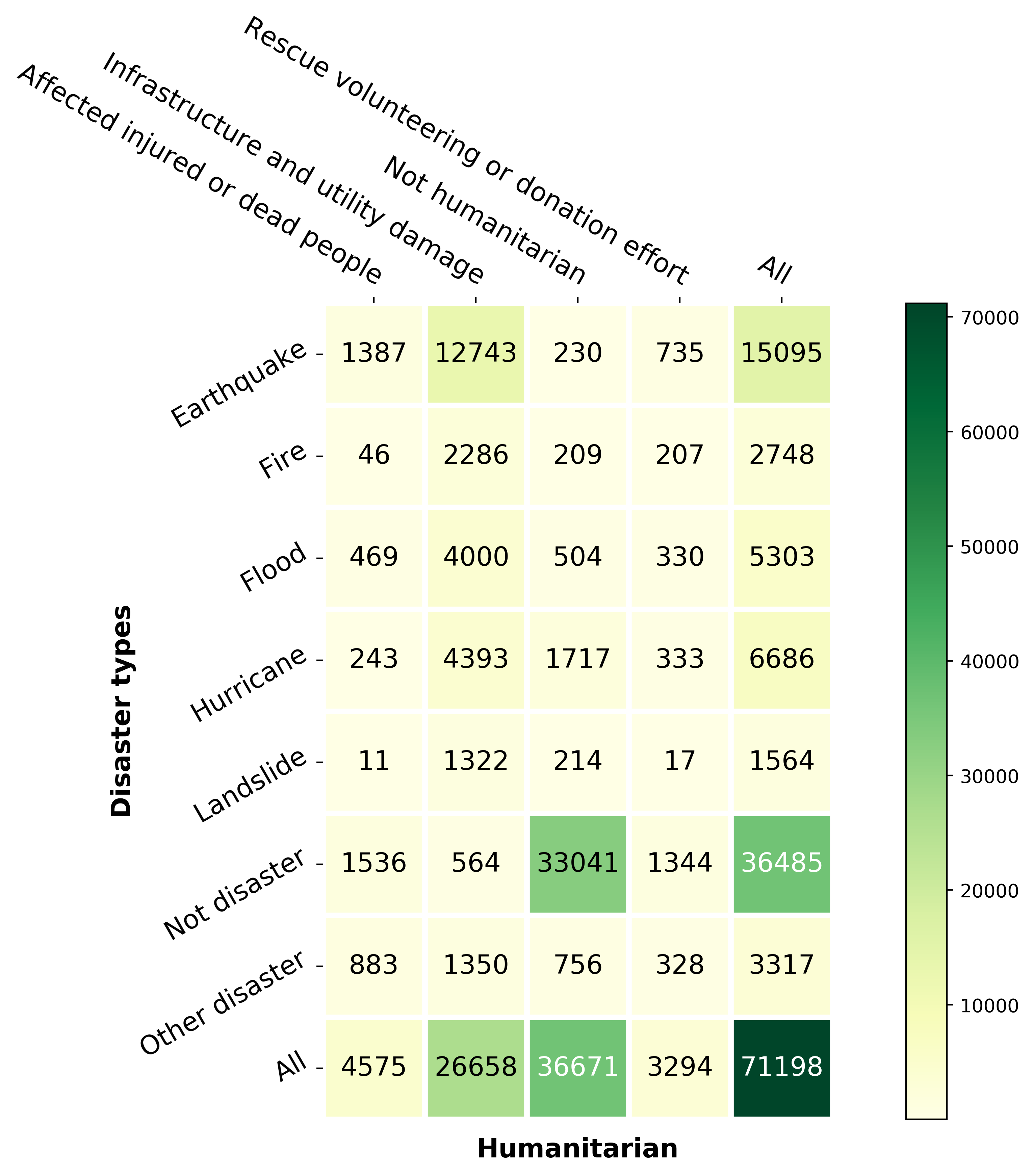}
        \caption{DS and hum.}
        \label{fig:contingency_table_dt_hum}    
    \end{subfigure}% 
    \begin{subfigure}[b]{0.45\textwidth}    
        \includegraphics[width=\textwidth]{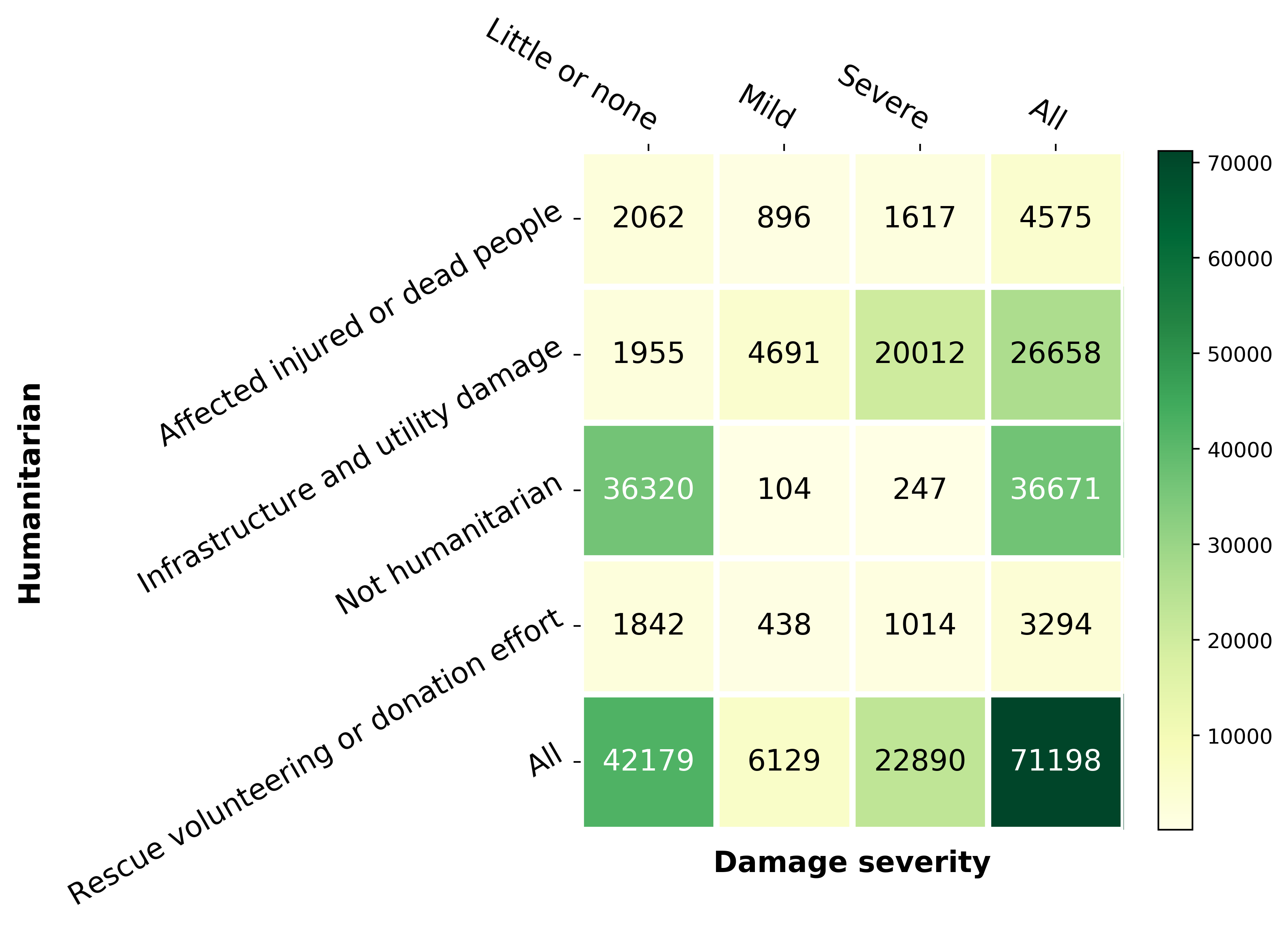}
        \caption{Hum and DS.}
        \label{fig:contingency_table_q4_q5}    
    \end{subfigure}
    \hfill
    \begin{subfigure}[b]{0.45\textwidth}    
        \includegraphics[width=\textwidth]{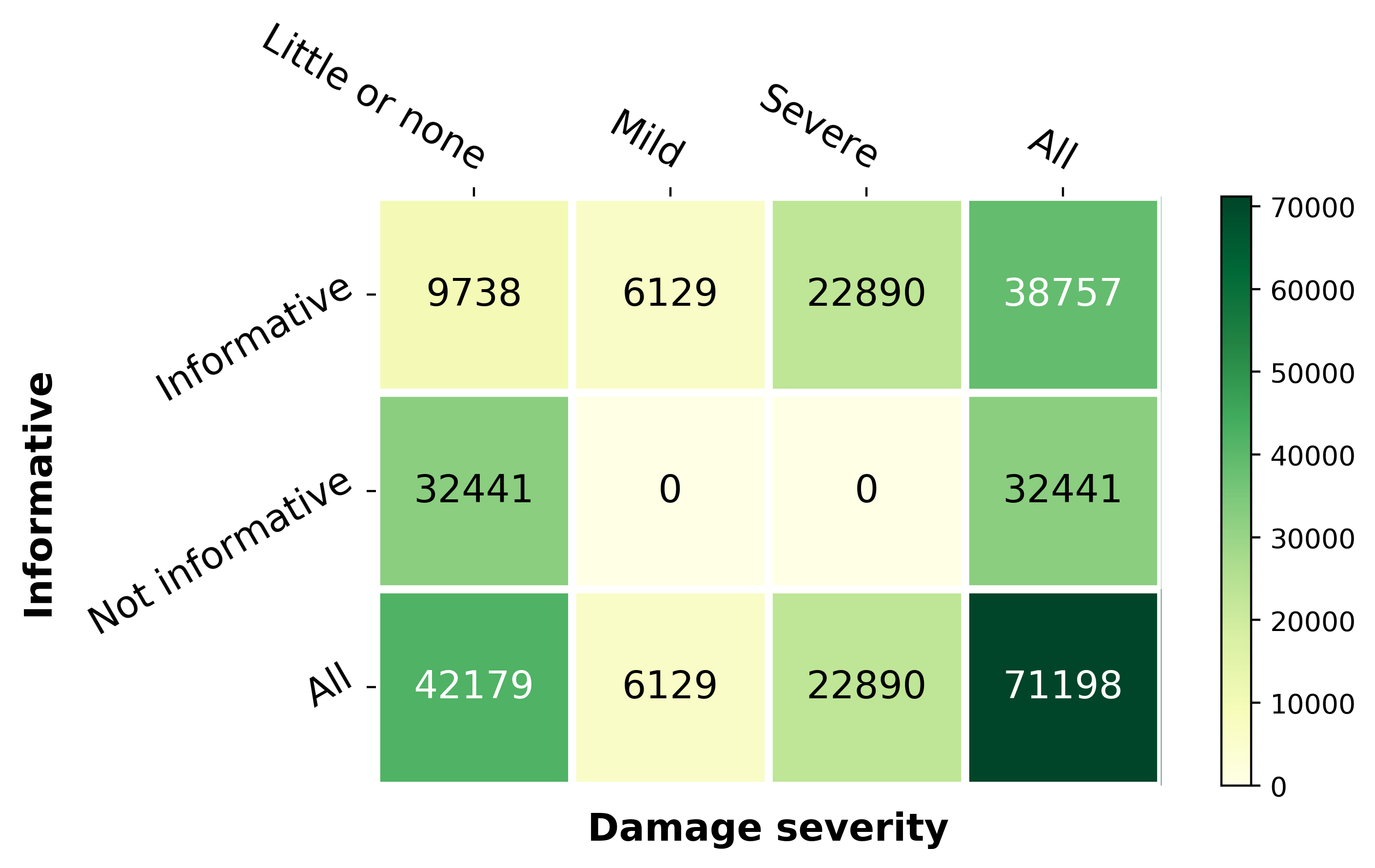}
        \caption{Info and DS.}
        \label{fig:contingency_table_q3_q5}    
    \end{subfigure}%
    \begin{subfigure}[b]{0.45\textwidth}    
        \includegraphics[width=\textwidth]{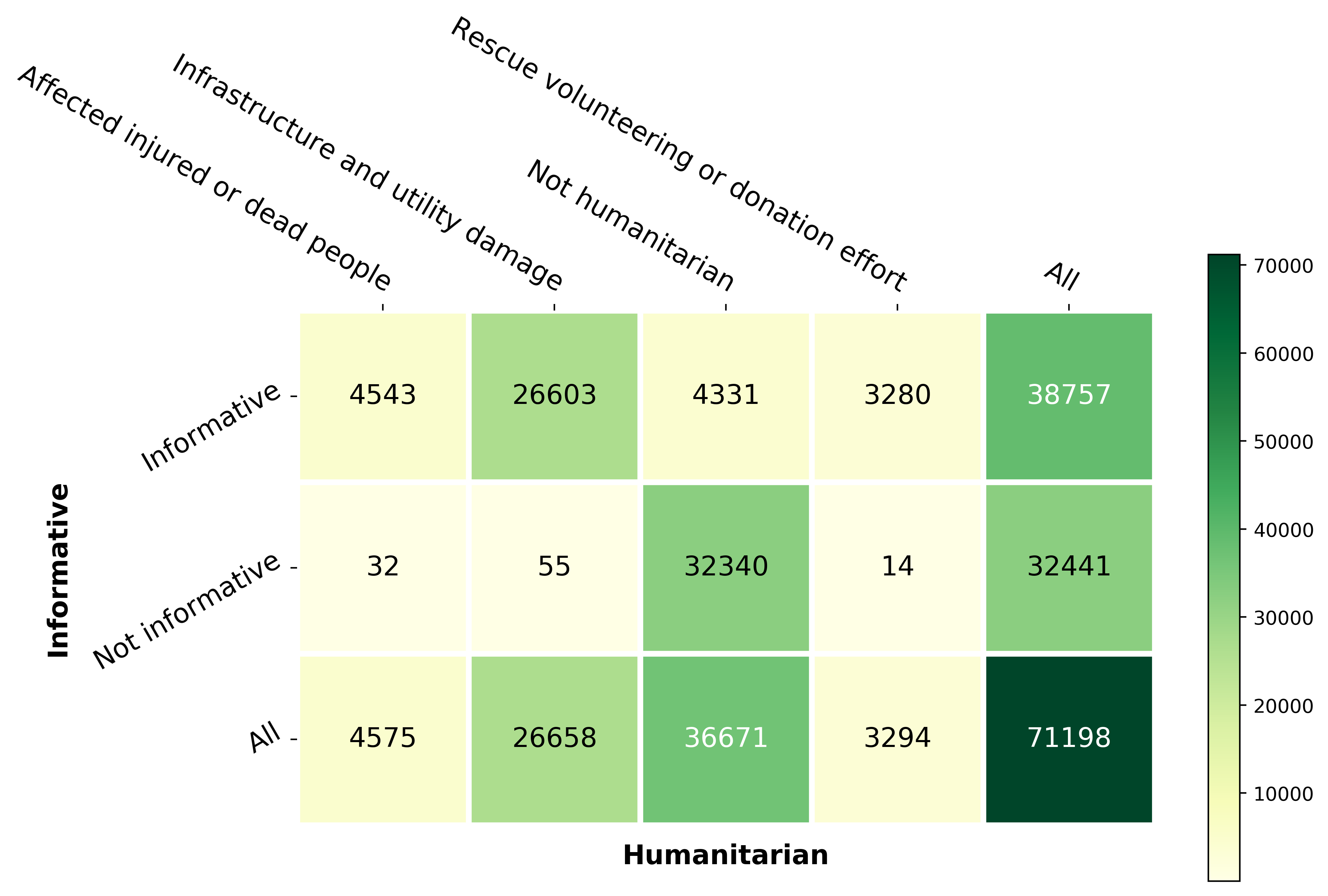}
        \caption{Info and hum.}
        \label{fig:contingency_table_info_ds}    
    \end{subfigure}    
    \caption{Contingency heatmaps for different pairs of tasks.}
    \label{fig:contingency_all}
\end{figure*}

\rev{
In Figure~\ref{fig:contingency_all}, we report class-wise relationship between tasks. It appears that there is an association between labels for different tasks. For example, for disaster types and informativeness tasks, as shown in the Figure \ref{fig:contingency_table_dt_info}, \textit{not disaster} and \textit{not informative} are highly related. A major part of \textit{not disaster} images are labeled as \textit{little or none} damages as shown in \ref{fig:contingency_table_dt_ds}. Our observations for other task combinations are quite similar for different label pairs.
}

\section{Error Analysis}
\label{sec:appendix:error_analysis}
In Tables \ref{tab:conf_mat_disaster_types}, \ref{tab:conf_mat_informative}, \ref{tab:conf_mat_humanitarian} and \ref{tab:conf_mat_damage_severity}, we report confusion matrices for different tasks with a comparison to single vs. multi-task settings. 

\begin{table}[]
\centering
\setlength{\tabcolsep}{2.5pt}
\scalebox{0.75}{
\begin{tabular}{@{}lrrrrrrrr@{}}
\toprule
\multicolumn{9}{c}{\textbf{Single-task}} \\ \midrule
\multicolumn{1}{@{}l}{\textbf{Label}} & \multicolumn{1}{r}{\textbf{Earthquake}} & \multicolumn{1}{r}{\textbf{Fire}} & \multicolumn{1}{r}{\textbf{Flood}} & \multicolumn{1}{r}{\textbf{Hurricane}} & \multicolumn{1}{r}{\textbf{Landslide}} & \multicolumn{1}{r}{\textbf{Not disaster}} & \multicolumn{1}{r}{\textbf{Other disaster}} & \multicolumn{1}{r@{}}{\textbf{Total}} \\ \midrule
Earthquake & 1482 & 22 & 14 & 66 & 38 & 156 & 17 & 1795 \\
Fire & 17 & 588 & 3 & 9 & 4 & 66 & 3 & 690 \\
Flood & 19 & 5 & 1061 & 64 & 20 & 145 & 1 & 1315 \\
Hurricane & 104 & 10 & 92 & 1025 & 29 & 234 & 24 & 1518 \\
Landslide & 27 & 3 & 7 & 13 & 260 & 21 & 0 & 331 \\
Not disaster & 122 & 53 & 142 & 241 & 28 & 8253 & 46 & 8885 \\
Other disaster & 237 & 71 & 39 & 144 & 39 & 407 & 217 & 1154 \\
Total & 2008 & 752 & 1358 & 1562 & 418 & 9282 & 308 & 15688 \\  \midrule
\multicolumn{9}{c}{\textbf{Multi-task}} \\  \midrule
% \multicolumn{1}{c}{\textbf{Label}} & \multicolumn{1}{c}{\textbf{Earthquake}} & \multicolumn{1}{c}{\textbf{Fire}} & \multicolumn{1}{c}{\textbf{Flood}} & \multicolumn{1}{c}{\textbf{Hurricane}} & \multicolumn{1}{c}{\textbf{Landslide}} & \multicolumn{1}{c}{\textbf{Not disaster}} & \multicolumn{1}{c}{\textbf{Other disaster}} & \multicolumn{1}{c}{\textbf{Total}} \\ \midrule
Earthquake & 1498 & 22 & 12 & 69 & 34 & 150 & 10 & 1795 \\
Fire & 21 & 589 & 3 & 11 & 3 & 60 & 3 & 690 \\
Flood & 24 & 6 & 1062 & 55 & 20 & 147 & 1 & 1315 \\
Hurricane & 151 & 16 & 112 & 956 & 41 & 232 & 10 & 1518 \\
Landslide & 30 & 4 & 5 & 16 & 250 & 26 & 0 & 331 \\
Not disaster & 130 & 76 & 121 & 243 & 34 & 8237 & 44 & 8885 \\
Other disaster & 272 & 82 & 38 & 135 & 32 & 415 & 180 & 1154 \\
Total & 2126 & 795 & 1353 & 1485 & 414 & 9267 & 248 & 15688 \\ \bottomrule
\end{tabular}
}
\caption{Confusion matrix for \textbf{disaster types} task using single vs. multitask learning with efficient-net (b1) model.}
\label{tab:conf_mat_disaster_types}
\end{table}

\begin{table}[]
\centering
\setlength{\tabcolsep}{2.5pt}
\scalebox{0.75}{
\begin{tabular}{@{}lrrr|rrr@{}}
\toprule
\multicolumn{1}{c}{\textbf{}} & \multicolumn{3}{c|}{\textbf{Single-task}} & \multicolumn{3}{c}{\textbf{Multi-task}} \\ \midrule
\multicolumn{1}{@{}l}{\textbf{Label}} & \multicolumn{1}{r}{\textbf{Informative}} & \multicolumn{1}{r}{\textbf{Not Informative}} & \multicolumn{1}{r|}{\textbf{Total}} & \multicolumn{1}{r}{\textbf{Informative}} & \multicolumn{1}{r}{\textbf{Not Informative}} & \multicolumn{1}{r@{}}{\textbf{Total}} \\ \midrule
Informative & 6256 & 950 & 7206 & 6489 & 717 & 7206 \\
Not Informative & 977 & 7505 & 8482 & 1076 & 7406 & 8482 \\
Total & 7233 & 8455 & 15688 & 7565 & 8123 & 15688 \\ \bottomrule
\end{tabular}
}
\caption{Confusion matrix for \textbf{informative} task using single vs. multitask learning with efficient-net (b1) model.}
\label{tab:conf_mat_informative}
\end{table}

\begin{table}[]
\centering
\setlength{\tabcolsep}{2.5pt}
\scalebox{0.75}{
\begin{tabular}{@{}lrrrrr@{}}
\toprule
\multicolumn{6}{c}{\textbf{Single-task}} \\ \midrule
Label & \multicolumn{1}{l}{Affected} & \multicolumn{1}{l}{Infra. damage} & \multicolumn{1}{l}{Not hum} & \multicolumn{1}{l}{Rescue} & \multicolumn{1}{l}{Total} \\ \midrule
Affected, injured, or dead people & 272 & 181 & 149 & 37 & 639 \\
Infrastructure and utility damage & 68 & 4445 & 630 & 81 & 5224 \\
Not humanitarian & 93 & 649 & 8219 & 184 & 9145 \\
Rescue volunteering or donation effort & 63 & 180 & 150 & 287 & 680 \\
Total & 496 & 5455 & 9148 & 589 & 15688 \\ \midrule
\multicolumn{6}{c}{\textbf{Multi-task}} \\ \midrule
% \multicolumn{1}{c}{\textbf{Label}} & \multicolumn{1}{c}{\textbf{Affected, injured, or dead people}} & \multicolumn{1}{c}{\textbf{Infrastructure and utility damage}} & \multicolumn{1}{c}{\textbf{Not humanitarian}} & \multicolumn{1}{c}{\textbf{Rescue volunteering or donation effort}} & \multicolumn{1}{c}{\textbf{Total}} \\ \midrule
Affected, injured, or dead people & 280 & 191 & 137 & 31 & 639 \\
Infrastructure and utility damage & 67 & 4588 & 522 & 47 & 5224 \\
Not humanitarian & 118 & 698 & 8155 & 174 & 9145 \\
Rescue volunteering or donation effort & 79 & 214 & 141 & 246 & 680 \\
Total & 544 & 5691 & 8955 & 498 & 15688 \\ \bottomrule
\end{tabular}
}
\caption{Confusion matrix for \textbf{humanitarian} task using single vs. multitask learning with efficient-net (b1) model.}
\label{tab:conf_mat_humanitarian}
\end{table}

\begin{table}[]
\centering
\setlength{\tabcolsep}{2.5pt}
\scalebox{0.75}{
\begin{tabular}{@{}lrrrr|rrrr@{}}
\toprule
\multicolumn{1}{c}{\textbf{}} & \multicolumn{4}{c|}{\textbf{Single-task}} & \multicolumn{4}{c}{\textbf{Muli-task}} \\ \midrule
\multicolumn{1}{@{}l}{\textbf{Label}} & \multicolumn{1}{r}{\textbf{Little or none}} & \multicolumn{1}{r}{\textbf{Mild}} & \multicolumn{1}{r}{\textbf{Severe}} & \multicolumn{1}{r|}{\textbf{Total}} & \multicolumn{1}{r}{\textbf{Little or none}} & \multicolumn{1}{r}{\textbf{Mild}} & \multicolumn{1}{r}{\textbf{Severe}} & \multicolumn{1}{r@{}}{\textbf{Total}} \\ \midrule
Litle or none & 9550 & 120 & 582 & 10252 & 9476 & 152 & 624 & 10252 \\
Mild & 563 & 149 & 815 & 1527 & 481 & 179 & 867 & 1527 \\
Severe & 530 & 83 & 3296 & 3909 & 453 & 109 & 3347 & 3909 \\
Total & 10643 & 352 & 4693 & 15688 & 10410 & 440 & 4838 & 15688 \\ \bottomrule
\end{tabular}
}
\caption{Confusion matrix for \textbf{damage severity} task using single vs. multitask learning with efficient-net (b1) model.}
\label{tab:conf_mat_damage_severity}
\end{table}

\section{The MEDIC Dataset}
\label{sec:appendix:medic_dataset}
The dataset can be downloaded from \url{https://crisisnlp.qcri.org/medic/index.html}.

\subsection{Data Format}
\label{ssec:appendix:medic_dataset_format}
The dataset format can be found in \url{https://crisisnlp.qcri.org/medic/index.html}.

\subsection{Terms of Use, Privacy and License}
\label{ssec:appendix:medic_dataset_license}
% Terms of use, privacy and License. 
The MEDIC dataset is published under CC BY-NC-SA 4.0 license, which means everyone can use this dataset for non-commercial research purpose: \url{https://creativecommons.org/licenses/by-nc/4.0/}. 

\subsection{Data Maintenance}
\label{ssec:appendix:medic_dataset_maintenance}
We provide data download link through \url{https://crisisnlp.qcri.org/medic/index.html}. We also host the dataset on Dataverse\footnote{\url{https://dataverse.org/}} for wider access. We will maintain the data for a long period of time and make sure dataset is accessible.

\subsection{Benchmark Code}
The benchmark code is available at: \url{https://github.com/firojalam/medic/}.

\subsection{Ethics Statement}
\label{ssec:appendix:medic_dataset_biases}
% https://neurips.cc/public/EthicsGuidelines

\subsubsection{Dataset Collection}
\label{sssec:appendix:medic_dataset_coll}
The dataset contains images from multiple sources such as Twitter, Google, Bing, Flickr, and Instagram. Twitter developer terms and conditions suggests that one can release 50K tweet objects\footnote{\url{http://developer.twitter.com/en/developer-terms/agreement-and-policy}} and here we only provide images not whole JSON objects. The total number of images from Twitter is less than 50,000. Hence, by releasing the data by maintaining such terms and conditions. From Google, Bing, Yahoo and Instagram images are publicly available. In addition, we also maintain licenses and cite prior work based upon we built our work.

\subsubsection{Potential Negative Societal Impacts}
\label{sssec:appendix:medic_dataset_misuse}
The dataset consists of images collected from social media and different search engines. We have given our best efforts to eliminate any adult content during data preparation and annotation. Hence, we believe that the presence of such content in the dataset might be very unlikely. Our annotation does not contain any identifiable information such as age, gender, or race. However, the images in the dataset have many faces and one might apply facial recognition to identify someone. Intervention with human moderation would be required in order to ensure this does not lead to any misuse. We also would like to highlight that the models’ prediction should be used carefully as the purpose of the models’ prediction is to facilitate its user, not to make any direct decision. Model designers also need to be careful for any adversarial attack that can lead to creation and spread of any mis/disinformation.

\subsubsection{Biases}
\label{sssec:appendix:medic_dataset_biases}
The datasets are not representative of a geolocation, user gender, age, race, so should not be used in analyses requiring a representative sample. Instead, the datasets are more suitable to be combined with existing datasets and used for training supervised machine learning models.

We also would like to highlight that some of the annotations are subjective, and we have clearly indicated in the text which of these are. Thus, it is inevitable that there would be biases in our dataset. Note that, we have very clear annotation instructions with examples in order to reduce such biases. 

\subsubsection{Intended Use}
\label{sssec:appendix:medic_dataset_intended_use}
The dataset can enable an analysis of image content for disaster response, which could be of interest to crisis responders humanitarian response organizations, and policymakers. There are only very few datasets available for multitask learning research. This dataset can significantly help towards this direction. Having a single model for multiple tasks can also foster Green AI.